\documentclass[10pt,twocolumn,letterpaper]{article}

\usepackage[pagenumbers]{cvpr} 
\usepackage[accsupp]{axessibility}
\usepackage{graphicx}
\usepackage{amsmath}
\usepackage{amssymb}
\usepackage{booktabs}
\usepackage{color}
\usepackage{bm}
\definecolor{mycitecolor}{rgb}{0, 0.4, 0.7}
\usepackage{makecell}
\usepackage{arydshln}
\usepackage[breaklinks=true,bookmarks=true,citecolor=mycitecolor,colorlinks,pagebackref=true]{hyperref}

\usepackage{booktabs}

\usepackage[capitalize]{cleveref}
\crefname{section}{Sec.}{Secs.}
\Crefname{section}{Section}{Sections}
\Crefname{table}{Table}{Tables}
\crefname{table}{Tab.}{Tabs.}

\def\cvprPaperID{1617} 
\def\confName{CVPR}
\def\confYear{2023}

\begin{document}

\title{Think Twice before Driving: \\
Towards Scalable Decoders
for End-to-End Autonomous Driving}


\author{
Xiaosong Jia$^{1,2}$,
Penghao Wu$^{2,3}$,
Li Chen$^{2}$,
Jiangwei Xie$^{2}$,
Conghui He$^2$,
Junchi Yan$^{1,2^\dagger}$,
Hongyang Li$^{2,1^\dagger}$ \\
[2mm]
$^1$~Shanghai Jiao Tong University \quad 
$^2$~Shanghai AI Laboratory \\
$^3$~University of California at San Diego \quad
\\ 
\normalsize{
$^\dagger$Correspondence authors}\\
\normalsize{
\url{https://github.com/OpenDriveLab/ThinkTwice}
}
}

\twocolumn[{%
\renewcommand\twocolumn[1][]{#1}%
\maketitle
\begin{center}
    \centering
    \captionsetup{type=figure}
    \vspace{-10pt}
    \includegraphics[width=.8\textwidth,height=5cm]{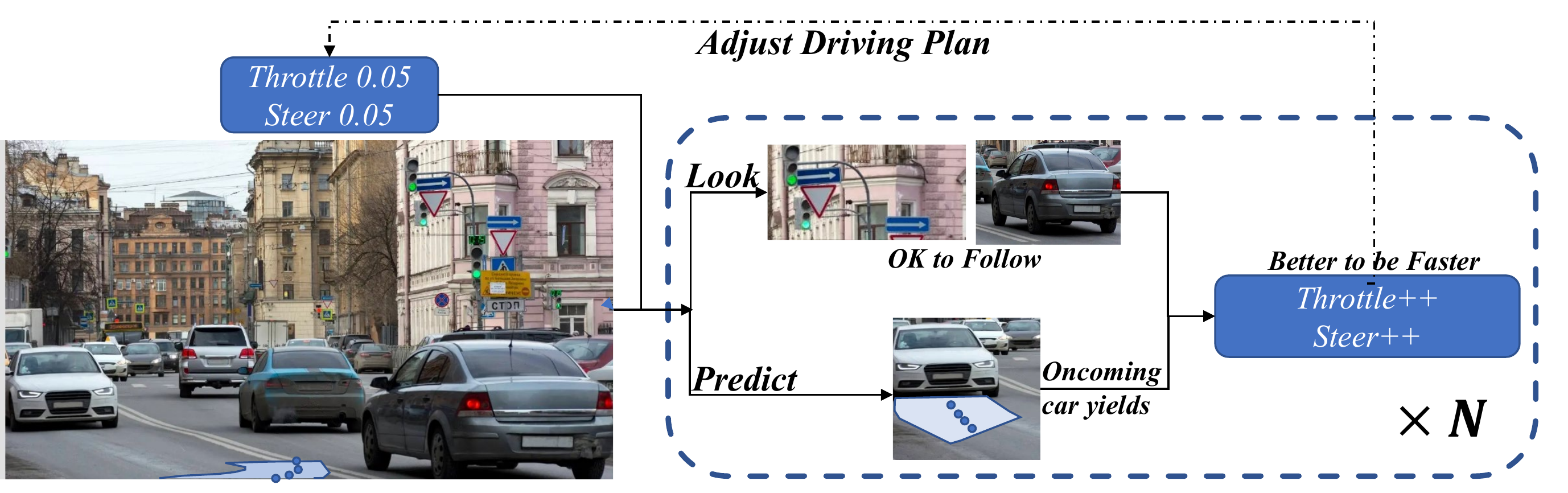}
    \vspace{-5pt}
    \captionof{figure}{We propose \textbf{ThinkTwice}, a scalable decoder paradigm that generates the future trajectory and action of the ego vehicle
     for end-to-end autonomous driving. Conditioned on the coarse action/trajectory, we propose the \emph{Look Module} to retrieve information from critical regions and the \emph{Prediction Module} to anticipate the outcome of the ego behavior. Taking features from the two modules as input, the coarse prediction is refined by predicting its offset from ground-truth. We could stack multiple such layers and scale up the capacity of the decoder with dense supervision and spatial-temporal prior.\label{fig:teaser}}
\end{center}}]

\begin{abstract}
\vspace{-10pt}
End-to-end autonomous driving has made impressive progress in recent years. 
Existing methods usually adopt the decoupled encoder-decoder paradigm, where the encoder extracts hidden features from raw sensor data, and the decoder outputs the ego-vehicle's future trajectories or actions.
Under such a paradigm, the encoder does not have access to the intended behavior of the ego agent, leaving the burden of finding out safety-critical regions from the massive receptive field and inferring about future situations to the decoder. Even worse, the decoder is usually composed of several simple multi-layer perceptrons (MLP) or GRUs while the encoder is delicately designed (e.g., a combination of heavy ResNets or Transformer).
Such an imbalanced resource-task division hampers the learning process.

In this work, we aim to alleviate the aforementioned problem by two principles:
(1) fully utilizing the capacity of the encoder;
(2) increasing the capacity of the decoder.
Concretely, we first predict a coarse-grained future position and action based on the encoder features. Then, conditioned on the position and action, the future scene is imagined to check the ramification if we drive accordingly.
We also retrieve the encoder features around the predicted coordinate to obtain fine-grained information about the safety-critical region. Finally, based on the predicted future and the retrieved salient feature, we refine the coarse-grained position and action by predicting its offset from ground-truth.
The above refinement module could be stacked in a cascaded fashion, which extends the capacity of the decoder with spatial-temporal prior knowledge about the conditioned future.
We conduct experiments on the CARLA simulator and achieve state-of-the-art performance in closed-loop benchmarks. Extensive ablation studies demonstrate the effectiveness of each proposed module.
%
   
\end{abstract}

\vspace{-5pt}
\section{Introduction}
\label{sec:intro}
\vspace{-4pt}

With the advance in deep learning, autonomous driving has attracted attention from both academia and industry. End-to-end autonomous driving~\cite{pomerleau1988alvinn,muller2005off} aims to build a fully differentiable learning system that is able to map the raw sensor input directly to a control signal or a future trajectory. Due to its efficiency and ability to avoid cumulative errors, impressive progress has been achieved in recent years~\cite{nvidia2016EndTE,codevilla2018cil,sauer2018cal,codevilla2019exploring,chen2020learning,chitta2021neat}.
State-of-the-art works~\cite{Prakash2021CVPR,chen2022lav,wu2022trajectoryguided,shao2022interfuser,hu2022model,wu2023policy} all adopt the encoder-decoder paradigm. The encoder module extracts information from raw sensor data (camera, LiDAR, Radar, \etc) and generates a representation feature. Taking the feature as input, the decoder directly predicts way-points or control signals. 

Under such a paradigm, the encoder does not have access to the intended behavior of the ego agent, which leaves the burden of finding out the safety-critical regions from the large perceptive field of massive sensor inputs and inferring about the future situations to the decoder.
For example, when the ego vehicle is at the intersection, if it decides to go straight, it should check the traffic light across the road, which might consist of only several pixels. If it decides to go right, then it should check whether there are any agents on its potential route and think about how they would react to the ego vehicle's action.
Even worse, the decoder is usually several simple multi-layer perceptrons (MLP) or GRUs while the encoder is a delicately designed combination of the heavy ResNet or Transformer.
Such unmatched resource-task division hampers the overall learning process.

%

To address the aforementioned issues, we design our new model based on two principles:
\begin{itemize}
\itemsep 0em
    \item \textbf{Fully utilize the capacity of the encoder}. Instead of leaving all future-related tasks to the decoder, we should reuse the features from the encoder conditioned on the predicted decision.
    \item \textbf{Extend the capacity of the decoder with dense supervision}. Instead of simply adding depth/width of MLP which would cause severe overfit, we should enlarge the encoder with prior structure and corresponding supervision so that it could capture the inherent driving logical reasoning.
\end{itemize}
To instantiate these two principles, we propose a cascaded decoder paradigm to predict the future action of the ego vehicle in a coarse-to-fine fashion as shown in Fig.~\ref{fig:teaser}.
Concretely, (i) We first adopt an MLP similar to classical approaches to generate the coarse future trajectory and action.
(ii) We then retrieve features around the predicted future location from the encoder and further feed them into several convolutional layers to obtain goal-related scene features (we denote the module as \emph{Look Module} and the feature as \emph{Look Feature}). This follows the intuition that human drivers would check their intended target to ensure safety and legitimacy.  
(iii) Inspired by the fact that human drivers would anticipate other agents' future motion to avoid possible collisions, we design a \emph{Prediction Module}, which takes the coarse action and features of the current scene as input and generates future scene representation features (denoted as \emph{Prediction Feature}).
 Considering the difficulty of obtaining supervision of the future scene representation conditioned on the predicted action during open-loop imitation learning, we adopt the teach-forcing technique~\cite{bengio2015scheduled}:  during training, we additionally feed samples with ground-truth action/trajectory into {Prediction Module} and supervise the corresponding \emph{Prediction Feature} with ground-truth future scene. As for the target of the supervision, we choose features from Roach~\cite{zhang2021roach}, an RL-based teacher network with privileged input, which contains decision-related information.
(iv) Based on the \emph{Look Feature} and \emph{Prediction Feature}, we predict the offset between the coarse prediction and ground-truth for refinement. The aforementioned process could be stacked cascadedly, which enlarges the capacity of the decoder with spatial-temporal prior knowledge about the conditioned future.

We conducted experiments on two competitive closed-loop autonomous driving benchmarks with CARLA~\cite{Dosovitskiy17} and achieved state-of-the-art performance. We also conducted extensive ablation studies to demonstrate the effectiveness of the components of the proposed method.

In summary, our work has three-fold contributions:
\begin{enumerate}
\setlength{\itemsep}{0pt}
    \item We propose a scalable decoder paradigm for end-to-end autonomous driving, which, to the best of our knowledge, is the first to emphasize the importance of enlarging the capacity of the decoder in this field.
    \item We devise a decoder module to look back to the safety critical areas and anticipate the future scene conditioned on the predicted action/trajectory, which injects spatial-temporal prior knowledge and dense supervisions into the training process.

    \item We demonstrate state-of-the-art performance on two competitive benchmarks and conduct extensive ablation studies to verify the effectiveness of the proposed module.
\end{enumerate}

We believe that the decoder (decision part) is equally important as the encoder (perception part) in end-to-end autonomous driving. We hope our exploration could inspire further efforts in this line of study for the community.

\begin{figure*}[tb!]
    \centering
    \includegraphics[width=\textwidth]{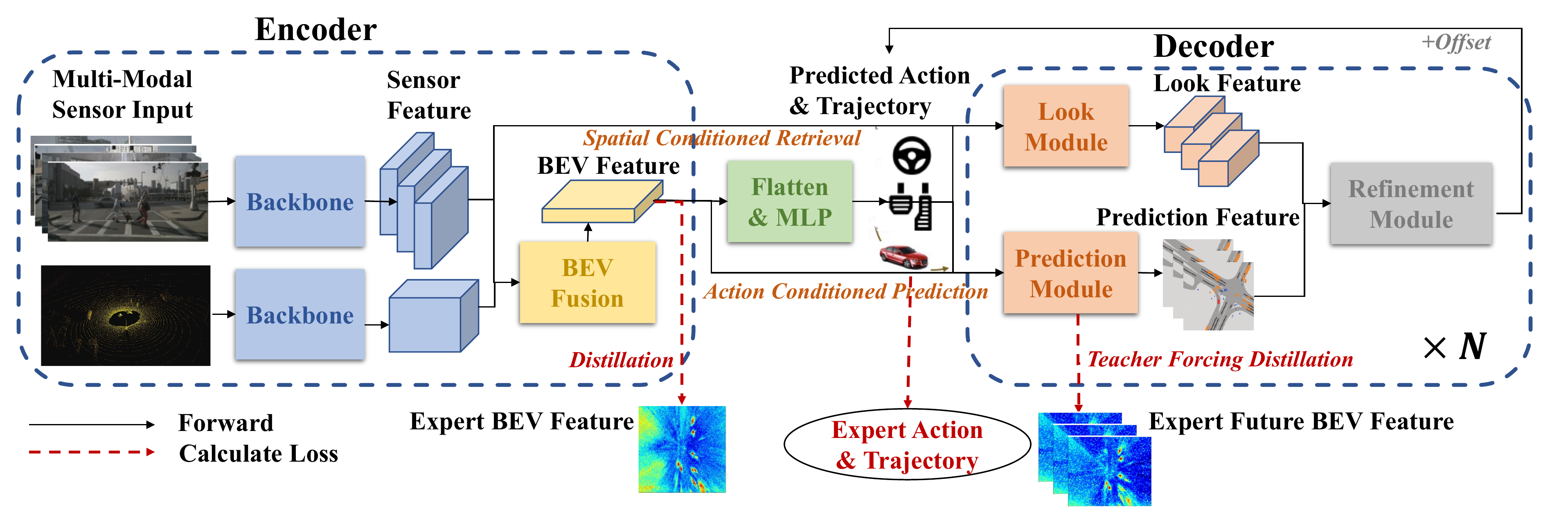}
    \vspace{-25pt}
    \caption{\textbf{Overall architecture of ThinkTwice.} (a) The encoder first processes raw data from different sensors with corresponding backbones and then fuses them into a  bird's-eye-view (BEV) representation. (b) With BEV features as input, the decoder first generates the coarse trajectory/action of the ego agent. Then,  the \emph{Look Module} retrieves sensor features around the predicted location - \emph{Look Feature} and the \emph{Prediction Module} anticipates the future BEV features conditioned on the predicted action - \emph{Prediction Feature}. (c) Based on these two features, the refinement module predicts the offset between the coarse output and the ground-truth (GT). \vspace{-5pt}    \label{fig:overall}}
\end{figure*}

\section{Related Work}
\vspace{-4pt}
\subsection{End-to-end Autonomous Driving}
\vspace{-4pt}
Unlike traditional modular autonomous driving frameworks, end-to-end methods, which predict actions based on sensor observations, have shown great potential. CIL \cite{codevilla2018cil} proposes a simple structure to directly map front-view image features to control signals based on navigation commands. Based on that, CILRS \cite{codevilla2019exploring} adds a speed prediction branch to alleviate the inertia problem. LBC \cite{chen2020learning} and Roach \cite{zhang2021roach} use learning-based privilege experts to better teach student models. Control and trajectory planning are combined in \cite{wu2022trajectoryguided} in a multi-task learning approach. Reinforcement learning methods \cite{toromanoff2020end,chekroun2021gri} use useful representations from pre-training tasks to accelerate the training process. NEAT \cite{chitta2021neat} decodes way-points and semantics in  bird’s-eye-view (BEV). Effective sensor fusion approaches to extract useful features from sensors for planning are explored in \cite{Prakash2021CVPR, chen2022lav, shao2022interfuser,zhang2022mmfn}. Model-based imitation learning is studied in \cite{hu2022model} to explicitly model the environment. 

Such methods have gained impressive performance in closed-loop evaluation. Nevertheless, most of them focus on the encoder part only, usually adopting a simple MLP or GRU-based decoder for final planning. In our work, we explore  increasing the capacity of the decoder and fully exploiting the capacity of the encoder simultaneously.

\subsection{BEV Representation for Autonomous Driving}
\vspace{-4pt}
Learning BEV representation for perception and planning tasks in autonomous driving is a heated topic~\cite{li2022delving} in both industry and academia. The BEV representation inherently preserves the spatial relationships on the ground plane, making it preferable for joint perception-planning and sensor fusion. 

Perception tasks in BEV including detection \cite{roddick2018orthographic,reading2021categorical,huang2021bevdet,liu2022petr,li2022bevdepth,li2022bevformer}, segmentation \cite{philion2020lift, hu2022st,li2022bevformer,xie2022m,zhou2022cross}, and lane detection \cite{garnett20193d,can2021structured,chen2022persformer} have been rapidly pushing the frontier of 3D vision for autonomous driving. For planning, BEV representation has also shown great potential, where the model could reason about the important geometry relationships. ChauffeurNet \cite{bansal2018chauffeurnet} renders the privileged  environment and routed information to BEV as input for planning. LBC \cite{chen2020learning}, LAV \cite{chen2022lav} and Roach \cite{zhang2021roach} train a strong expert based on privileged BEV representation as input. The cost-maps for planning can also be constructed or learned based on BEV representations \cite{sadat2020perceive, zeng2019end, zeng2020dsdnet, cui2021lookout, casas2021mp3, hu2022st,uniad}. NEAT \cite{chitta2021neat}  updates an attention map iteratively to aggregate features and decode them to semantic categories and way-points in BEV. However, NEAT does not explicitly convert image features in perspective view to BEV, and it only uses a simple MLP to directly decode results from the aggregated feature. In our work, after explicitly projecting and  aligning the image feature with LiDAR feature in BEV, we retrieve BEV features in critical regions and anticipate the future scene to iteratively refine the planning outputs.

\subsection{Coarse-to-fine Strategy}
\vspace{-4pt}
The coarse-to-fine strategy has been widely studied and used in the field of computer vision. Typical two-stage methods for 2D detection \cite{girshick14CVPRRCNN,girshick2015fast,renNIPS15fasterrcnn} and 3D detection \cite{shi2019pointrcnn,yang2019std,li2021lidar,shi2020pv,deng2021voxel,yin2021center} usually first propose coarse region proposals and then extract features based on the proposals to generate the refined final predictions. The coarse-to-fine approach also gains great success in tasks like optical flow estimation \cite{fischer2015flownet,ilg2017flownet,sun2018pwc,teed2020raft}, salient object detection \cite{7780449,deng18r,wangiccv17,Qin_2019_CVPR}, and trajectory prediction \cite{mangalam2021goals,jia2021ide,pmlr-v164-jia22a,jia2022towards,shi2022motion}.

As for planning in autonomous driving, LAV \cite{chen2022lav} also iteratively refines the predicted way-points. However, their refinement is only based on the original feature with a simple RNN. Our method uses the coarse prediction to retrieve features in critical regions and anticipate the future to better refine the coarse prediction.

\section{Approach}\label{sec:approach}
\vspace{-4pt}

The proposed ThinkTwice is an end-to-end autonomous driving framework consisting of an encoder to transform the raw sensor data into a representation vector, and a decoder to generate future trajectories or actions of the ego agent based on the representation vector. The overall architecture is shown in Fig.~\ref{fig:overall}. 

\subsection{BEV Encoder}
\vspace{-4pt}
In this work, we consider two commonly used sensors in autonomous driving: cameras and LiDAR. 
To fuse their information, we first transform the raw sensor data into bird's-eye-view (BEV) features respectively, and then directly concatenate BEV features since they have already been aligned in space. 

For \textbf{camera} inputs - RGB images from multiple views, we first use an image backbone (such as ResNet~\cite{he2016deep}) on each image to obtain its compact feature map.
To transform 2D images into BEV space, LSS~\cite{philion2020lift} is adopted: We first predict the discrete depth distribution of each pixel and scatter each pixel into discrete points along the camera ray, where the feature at each point is the product of its predicted depth and corresponding pixel feature. For each grid in BEV, we aggregate features from those points within the grid by Frustum Pooling\footnote{Please refer to \cite{philion2020lift} for details.}.
In this way, we could aggregate images from an arbitrary number of cameras into one $C\times B_H \times B_W$ feature map, where $C$ is the hidden dimension, $B_H$ and $ B_W$ are the height and width of the BEV grid.
Further, to introduce temporal cues, we aggregate the previous BEV of historical images by transforming it to current egocentric coordinate system according to the relative ego movement. Previous and current feature maps are thus spatially aligned and we could simply concatenate them to obtain the final BEV feature.
Additionally, we found that (i) ground-truth supervision for the depth prediction module is important, which aligns with the finding in the object detection field~\cite{li2022bevdepth}. 
(ii) When scattering the image features, it is beneficial to add a semantic segmentation module and scatter the predicted semantic scores as well. We conjecture that it increases the generalization ability of the end-to-end model by filtering out the unrelated texture information.\footnote{We give ablations to support these 
claims in the experiment section.}

For \textbf{LiDAR} inputs - point clouds, we employ the popular SECOND~\cite{yan2018second} which applies sparse 3D-convolution on the voxelized point clouds~\cite{zhou2018voxelnet}. Its final output is also a BEV feature map with a size of $C\times B_H \times B_W$. To utilize temporal information, similar to existing works in object detection field~\cite{Hu2022AFDetV2RT,yin2021center,sun2021rsn}, we concatenate aligned point clouds from multiple-frames with an additional channel to indicate the time-step.

When \textbf{fusing} the two BEV feature maps, we simply concatenate them into one and process it by a series of 2D convolutional layers. Since actions are the only direct supervision  in end-to-end autonomous driving which are too sparse for the high-dimensional multi-sensor input, we provide extra feature-level supervision for the BEV feature map. Specifically, we use middle BEV feature maps from Roach~\cite{zhang2021roach} as the target, an RL-based teacher network with privileged input, which takes rasterized BEV surrounding environment as privileged input and achieves decent performance with several convolutional layers. Note that any learnable expert model with a BEV feature representation could be adopted here such as~\cite{chen2020learning,Renz2022CORL} and we adopt Roach here due to its robustness from RL training.  By letting the middle BEV feature maps of the student network (\ie, the encoder of ThinkTwice) be similar to the teacher network's, each BEV grid obtains dense supervision regarding decision-related information. In the experiment section, we empirically show that this supervision is necessary and is better than the commonly used BEV segmentation supervision signals in previous SOTA works~\cite{Prakash2021CVPR,chen2022lav,hu2022model}.

\subsection{Decoder}
\vspace{-4pt}
\subsubsection{Coarse Prediction Module}
\vspace{-4pt}
Recall that from the encoder part, we have obtained a BEV feature map $\bm{H}_{\text{BEV}}$ with the shape $C\times B_H \times B_W$.  We use 2D convolutional layers to downsample it and then flatten the small feature map into a 1D vector $\bm{H}_{\text{env}}$, which contains information about surrounding environments from cameras and LiDAR.
For routing information including the target point, high-level command (go straight, turn left, turn left, \etc), and current speed, we use an MLP to encode them into a compact vector $\bm{H}_{\text{mst}}$, similar to~\cite{chen2022lav,zhang2021roach,wu2022trajectoryguided}. 
With $\bm{H}_{\text{env}}$ and $\bm{H}_{\text{mst}}$ as input, we use another MLP to predict the ego vehicle's future action $\textbf{Ctrl}_{0}$ and trajectories $\textbf{Traj}_{0}$, where $0$ denotes it is the initialization of the prediction.
Note that this module follows the common practice in existing works: \emph{flatten + MLP}, which ignores the spatial-temporal association between the prediction and current observation. In the following part, we propose \emph{Look Module} and \emph{Prediction Module} to extend the capacity of the decoder with the aforementioned prior knowledge and dense supervision.

\subsubsection{Look Module}
\vspace{-4pt}

\begin{figure}[tb!]
    \centering
    \includegraphics[width=0.45\textwidth]{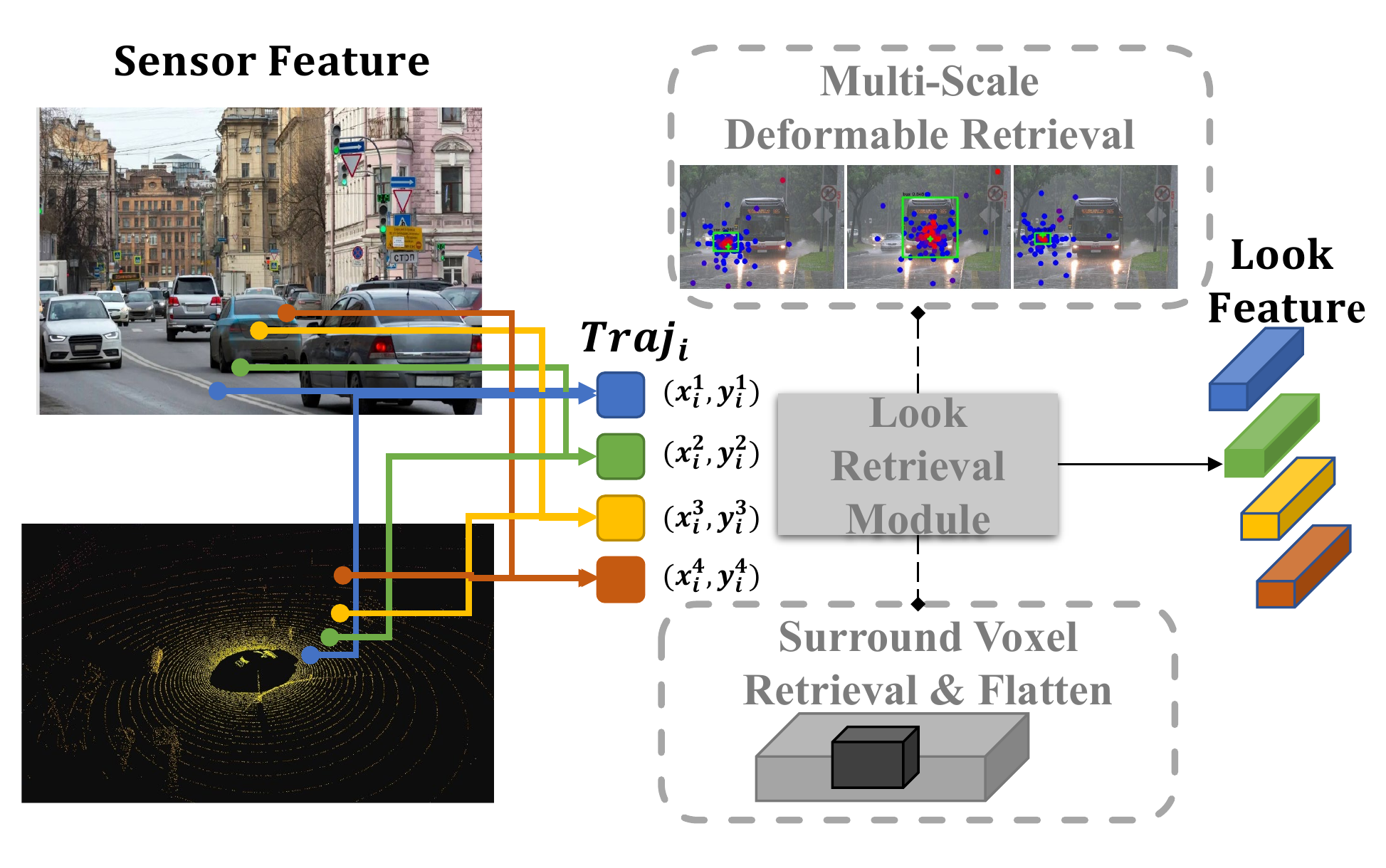}
    \vspace{-10pt}
    \caption{\textbf{{Overall architecture of Look Module.}} \vspace{-10pt}}
\label{fig:look}
\end{figure}

The intuition behind \emph{Look Module} is that human drivers would check their target location to make sure there are no collisions with other agents and no violations of traffic rules  before they actually go there, which is demonstrated to be effective in the trajectory prediction field~\cite{marchetti2020mantra}. 
Thus, with the predicted trajectory $\textbf{Traj}_i$ from the last layer with the shape $T\times 2$ where $T$ is the prediction horizon and $2$ represents (x, y), we retrieve sensor features based on the coordinate of $\textbf{Traj}_i$. The overall architecture is shown in Fig.~\ref{fig:look}.

For cameras, we project the coordinate back into the image plane with the cameras' extrinsics and intrinsics. 
Considering information on one pixel is limited and there might be errors during the projection, we adopt multi-scale deformable attention~\cite{zhu2020deformable} to aggregate information:
\begin{equation}
    \bm{H}^\text{img-look}_{i+1} = \text{DeformAttn}(\bm{H}_\text{Img};\textbf{Traj}_i; 
    \bm{H}_\text{env}, \bm{H}_\text{mst}),
\end{equation}
where $\bm{H}_\text{Img}$ is the multi-scale image feature maps; $\textbf{Traj}_i$ serves as the reference point of the deformable attention; $\bm{H}_\text{env}$ and $\bm{H}_\text{mst}$ serve as the query of the attention. 

For LiDAR, since it is already in the form of voxel features and the coordinate could be directly used, we simply retrieve the surrounding voxels of each coordinate in $\textbf{Traj}_i$ and flatten them followed by an MLP to obtain $\bm{H}^\text{lidar-look}_{i+1}$. Finally, we concatenate $\bm{H}^\text{img-look}_{i+1}$ and $\bm{H}^\text{lidar-look}_{i+1}$, and use an MLP to get the look feature $\bm{H}^\text{look}_{i+1}$. $H_\text{env}$ is also updated by $\bm{H}^\text{look}_{i+1}$ with another MLP.

By far, we reuse the representation power of the encoder and inject sample specific spatial prior, \ie, intended location, into the features, which makes the model easier to be optimized and could lead to better generalization ability~\cite{carion2020end,zhang2022dino}. 

\subsubsection{Prediction Module}
\vspace{-4pt}

The intuition behind the \emph{Prediction Module} is that human drivers would anticipate how surrounding agents would react to their action and check whether there would be collisions before they actually execute any action, \ie, action-conditioned prediction~\cite{malla2020titan}. 
Thus, to model the intended action-conditioned future of the scene, we use a spatial-GRU, which simply replaces linear layers in GRU~\cite{cho2014properties} with 2D convolutional layers. 
It takes the current BEV feature as its initial states and at each time-step takes the predicted coarse action $\textbf{Ctrl}_{i}^{t}$  and trajectories $\textbf{Traj}_{i}^{t}$ from the last layer  as input. 
We denote its output as $\bm{H}_\text{predict}^{i+1}$ with the shape $T \times B_{H} \times B_{W} \times C$ where $T$ is the prediction horizon,  $B_{H}$ and $B_{W}$ are the height and width of BEV grid, and $C$ is the hidden dimension.

\begin{figure}[tb!]
    \centering
    \includegraphics[width=0.45\textwidth]{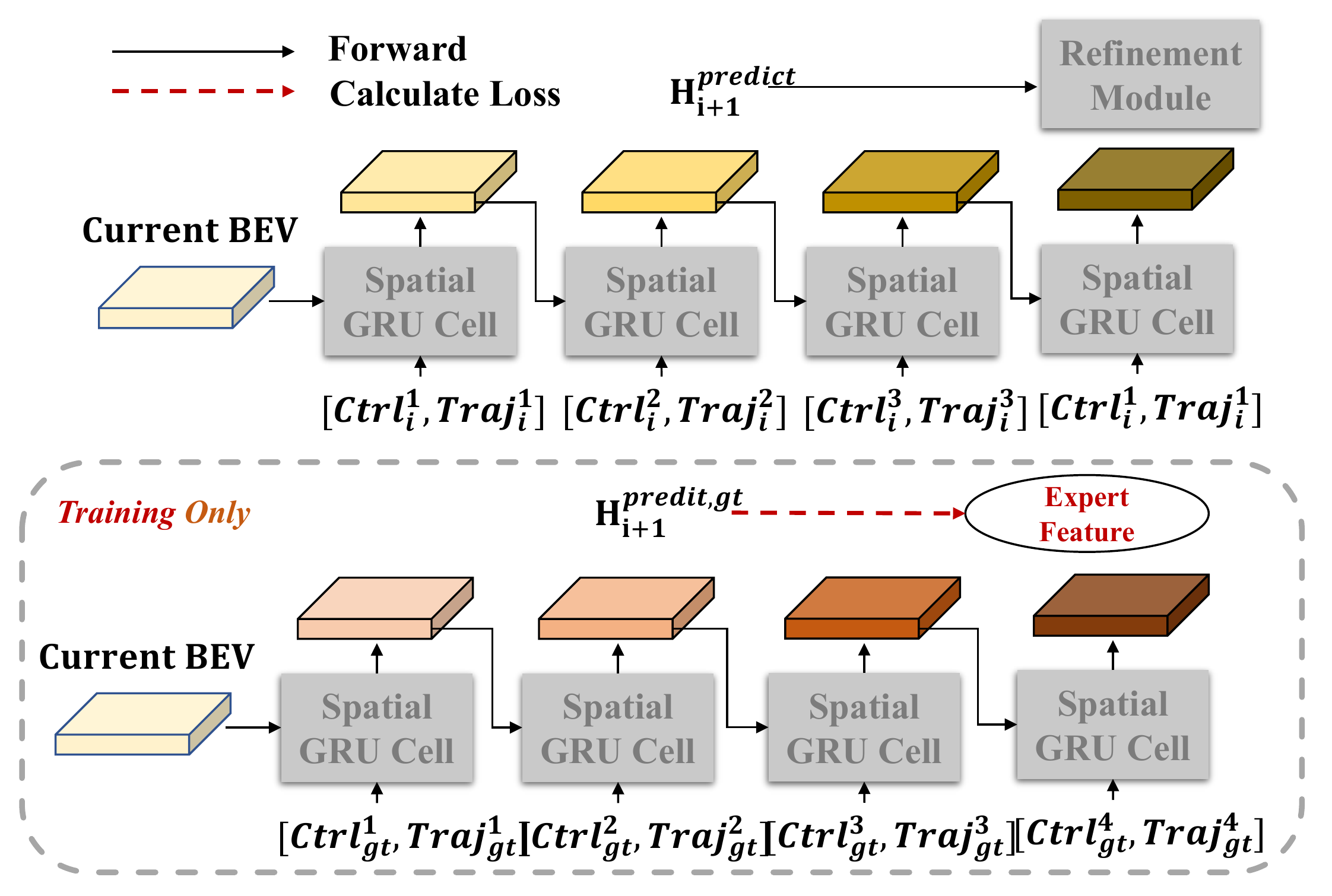}
    \vspace{-5pt}
    \caption{\textbf{{Overall architecture of Prediction Module.}}
    \vspace{-5pt}
    }
    \label{fig:predict}
\end{figure}

To provide supervision for $\bm{H}_\text{predict}^{i+1}$, we need to know the actual future scene when the ego vehicle takes the action $\textbf{Ctrl}_{i}$, which is difficult during the open-loop imitation learning process.
To deal with this issue, inspired by the teacher forcing~\cite{bengio2015scheduled} technique in NLP field, during training, we feed a set of extra inputs to the spatial GRU: the  current BEV feature with the ground-truth action and trajectory $\textbf{Ctrl}^{gt}$ and $\textbf{Traj}^{gt}$. 
Denoting its corresponding output as $\bm{H}^\text{predict,gt}_{i+1}$, we can supervise this hidden feature with the collected future scene.
Here, we choose the Roach BEV feature as the target.
Its overall structure is given in Fig.~\ref{fig:predict}.

\subsubsection{Refinement Module}
\vspace{-4pt}
With the \emph{Look Featue} $\bm{H}^\text{look}_{i+1}$ and \emph{Prediction Feature} $\bm{H}^\text{predict}_{i+1}$, the refinement module utilizes them to adjust the predicted action $\textbf{Ctrl}_{i}$ and trajectories $\textbf{Traj}_{i}$ from the last layer by:
\begin{equation}
\begin{split}
    \mathcal{O}_{i+1}^{\text{ctrl}},
    \mathcal{O}_{i+1}^{\text{traj}} =  \textbf{MLP} &  ([\bm{H}^\text{look}_{i+1}; \bm{H}^\text{predict}_{i+1};\textbf{Ctrl}_{i}; \\& \textbf{Traj}_{i};\bm{H}_\text{env};\bm{H}_\text{mst}]),
\end{split}
\end{equation}
where  $\mathcal{O}_{i+1}^{\text{ctrl}}$ and $\mathcal{O}_{i+1}^{\text{traj}}$ are the predicted offsets of action and trajectory between coarse prediction and ground-truth respectively. They are supervised as follows:
\begin{equation}
\begin{split}
    & \mathcal{O}_{i+1}^{\text{ctrl}} = \mathcal{L}(\textbf{Ctrl}_{i}, \textbf{Ctrl}^{gt}), \\
    & \mathcal{O}_{i+1}^{\text{traj}} = \mathcal{L}(\textbf{Traj}_{i}, \textbf{Traj}^{gt}),
\end{split}
\end{equation}
where $\mathcal{L}$ represents the loss function which is the commonly used Smooth L1 loss. Finally, we update the predicted action and trajectory by:
\begin{equation}
\begin{split}
    &  \textbf{Ctrl}_{i+1} = \textbf{Ctrl}_{i} + \mathcal{O}_{i+1}^{\text{ctrl}}, \\
    & \textbf{Traj}_{i+1} = \textbf{Traj}_{i} + \mathcal{O}_{i+1}^{\text{Traj}}.
\end{split}
\end{equation}

By then, we finish one round of refinement. Similar to DETR-based methods~\cite{carion2020end,zhang2022dino} in the general vision domain, the proposed decoder module could be stacked in cascade and we observe notable performance gain with multi-layers.

\begin{table*}[!t]
\centering
\small
\scalebox{0.9}{
\begin{tabular}{lccccccc}
\toprule
Method & Encoder & Decoder      & Modality  & Extra Supervision        & DS$\uparrow$            & RC$\uparrow$           & IS$\uparrow$           \\ \midrule
CILRS~\cite{codevilla2019exploring} & ResNet + Flatten  &   MLP     &  C1 & None         & 7.8$\pm$0.3    & 10.3$\pm$0.0  & 0.75$\pm$0.05 \\
LBC~\cite{chen2020learning} & ResNet + Flatten  &   MLP     & C3 & Expert  & 12.3$\pm$2.0   & 31.9$\pm$2.2  & 0.66$\pm$0.02 \\
Transfuser~\cite{Chitta2022PAMI} & Fusion via Transformer & GRU & C3L1 & Dep+Seg+Map+Box & 31.0$\pm$3.6   & 47.5$\pm$5.3  & \textbf{0.77}$\pm$0.04 \\
Roach~\cite{zhang2021roach} & ResNet + Flatten & MLP      & C1 & Expert       & 41.6$\pm$1.8   & \textbf{96.4}$\pm$2.1  & 0.43$\pm$0.03 \\
LAV~\cite{chen2022lav}    & PointPaiting & Multi-layer GRUs     & C4L1 & Expert+Seg+Map+Box        & 46.5$\pm$2.3   & 69.8$\pm$2.3  & 0.73$\pm$0.02 \\
TCP~\cite{wu2022trajectoryguided}   & ResNet + Flatten & GRU      & C1  & Expert    & 57.2$\pm$1.5  & 80.4$\pm$1.5 & 0.73$\pm$0.02 \\
\textbf{ThinkTwice}    & Geometric Fusion in BEV &     Look-Predict-Refine &  C4L1  & Expert+Dep+Seg+Map    &  \textbf{65.0}$\pm$1.7  & 95.5$\pm$2.0  & 0.69$\pm$0.05 \\ \midrule
MILE*\dag~\cite{hu2022model}  & ResNet + Flatten & GRU     & C1  & Map+Box    & 61.1$\pm$3.2   & \textbf{97.4}$\pm$0.8  & 0.63$\pm$0.03 \\
Interfuser*~\cite{shao2022interfuser} & Fusion via Transformer & Transformer + GRU  & C3L1   & Map+Box     & 68.3$\pm$1.9 & 95.0$\pm$2.9 & -            \\
\textbf{ThinkTwice}*   & Geometric Fusion in BEV &     Look-Predict-Refine     & C4L1 & Expert+Dep+Seg+Map              & \textbf{70.9}$\pm$3.4 & 95.5$\pm$2.6 & \textbf{0.75}$\pm$0.05 \\ \bottomrule
\end{tabular}}
\vspace{-5pt}
\caption{\textbf{Performance on Town05 Long benchmark.} $\uparrow$ means the higher the better. * denotes using extra data. $\dag$ denotes no scenarios are involved in the evaluation, which is a much easier benchmark. For \emph{Modality}, C denotes the camera sensor and L denotes the LiDAR sensor. \textit{Extra Supervision} refers to labels required to train their student model besides actions and states of the ego vehicle. \emph{Expert} denotes the distillation from privileged agents' outputs or features. \emph{Seg} and \emph{Depth} denotes the depth and semantic segmentation labels of the 2D images. \emph{Box} denotes the bounding box of surrounding agents. \vspace{-5pt} \label{tab:town05-long}}
\end{table*}

\subsection{Supervision Signals}
\vspace{-4pt}
In end-to-end autonomous driving, since the direct supervision - the action (2 float number) or trajectories (T*2 float number) is rather sparse compared to the multi-modal multi-view inputs, to avoid overfitting and increase the generalization ability, it is a common practice in SOTA works~\cite{Prakash2021CVPR,chen2022lav,wu2022trajectoryguided,shao2022interfuser,hu2022model} to apply multiple auxiliary supervisions. In ThinkTwice, we apply dense supervision on both the encoder and decoder and we summarize as follows:

\smallskip
\noindent\textbf{Image Depth \& Segmentation}: Compared to LiDARs, images contain more semantics yet with lots of abundance~\cite{liu2022bevfusion}. and it is more difficult to be applied in autonomous driving due to the discrepancy between 2D images and the 3D world. To address this, we apply depth supervision for more accurate BEV projection, and project semantic feature maps to BEV along with pixel feature maps so that the influence of texture could be mitigated.

\smallskip
\noindent\textbf{Feature Distillation}: Existing works have shown that with privileged input (ground-truth agent location, lanes, traffic light states, \etc), a learning-based teacher network could achieve decent performance~\cite{zhang2021roach,Renz2022CORL}. Compared to BEV segmentation - an auxiliary task, distilling the teacher network's middle feature maps provides more direct supervision and is more related to the major task - driving~\cite{zhang2021roach,wu2022trajectoryguided}. Here, we apply distillations on both the current BEV and the predicted future BEV (with teacher forcing).

\smallskip
\noindent\textbf{Auxiliary Tasks}: We use $\bm{H}_{\text{env}}$ to predict the ego vehicle’s current speed which is helpful to mitigate the inertia problem~\cite{zhang2021roach,wu2022trajectoryguided}. We use $\bm{H}_{\text{env}}$ and $\bm{H}_{\text{mst}}$ to estimate the value function of current states supervised by Roach - an RL teacher network. Similarly, for future scene representation generated by teacher forcing, we predict their speed and value as well.

\smallskip
\noindent\textbf{Direct Supervision}: Since we need coordinates to retrieve sensor features and actions to predict the conditioned future, our model outputs the ego vehicle's future action and trajectory together at each layer in a coarse-to-fine fashion. 

\smallskip
To this end, the encoder receives dense supervision from depth, segmentation, and feature distillation, while the decoder receives dense supervision from multi-layer direct supervision and future feature map supervision in the prediction module.

\section{Experiments}
\vspace{-4pt}

\subsection{Benchmark}
\vspace{-4pt}
We use CARLA~\cite{Dosovitskiy17} as the simulator to conduct closed-loop autonomous driving evaluations. 
We conduct experiments on two widely used benchmarks, \textbf{Town05 Long} and \textbf{Longest6}.
Each benchmark contains several routes and each route is defined by a sequence of sparse navigation points together with high-level commands (straight, turn left/right, lane changing, and lane following). 
The closed-loop driving task requires the autonomous agent to drive toward the destination point. It is designed to simulate realistic traffic situations and includes different challenging scenarios such as obstacle avoidance, crossing an unsignalized intersection, and sudden control loss.

\subsection{Data Collection}
\vspace{-4pt}
ThinkTwice is an imitation learning framework that requires an expert to collect driving logs: a sequence of vehicle states and sensor data.
Here, we use 4 cameras (front, left, right, back), one LiDAR, IMU, GPS, and speedometer.
We adopt Roach~\cite{zhang2021roach}, an RL-based network, as our expert similar to~\cite{wu2022trajectoryguided,hu2022model} due to the strong supervision it could provide. We collect data in 2 Hz on town01, town03, town04, and town06. We collect 189K frames in total which is similar to~\cite{chen2022lav,Chitta2022PAMI,wu2022trajectoryguided} to conduct most experiments and ablation studies.
To match the size of the dataset and number of seen towns with concurrent works~\cite{shao2022interfuser} (3M, 8 towns, 2Hz) and \cite{hu2022model} (2.9M, 4 towns, 25Hz), we additionally collect a dataset of 2M~\footnote{We stop collecting when our hard-disk is full.} with all 8 towns where we would only use to compare with them and we denote this setting with *.

\subsection{Metrics}
\vspace{-4pt}
We use the official metrics of CARLA Leaderboard: \textbf{Route Completion (RC)} is the percentage of the route completed by the autonomous agent. \textbf{Infraction Score (IS)} measures the number of infractions made along the route, with pedestrians, vehicles, road layouts, red lights, and \textit{etc}.\footnote{Please refer to \url{https://leaderboard.carla.org/get_started_v1/} for details.} \textbf{Driving Score (DS)} is the \emph{main metric} which is the product of Route Completion and Infraction Score.

\begin{table}[t!]
\centering
\begin{tabular}{lccc}
\toprule
Method      & DS$\uparrow$            & RC$\uparrow$            & IS$\uparrow$            \\ \midrule
WOR~\cite{chen2021learning}         & 23.6          & 52.3          & 0.59          \\ 
LAV~\cite{chen2022lav}         & 34.2          & 73.5          & 0.53          \\ 
Transfuser~\cite{Prakash2021CVPR,Chitta2022PAMI}  & 56.7          & \textbf{92.3} & 0.62          \\ 
\textbf{ThinkTwice}  & 61.3          & 73.0          & 0.81          \\ 
\textbf{ThinkTwice}* & \textbf{66.7} & 77.2          & \textbf{0.84} \\ \bottomrule
\end{tabular}
\vspace{-5pt}
\caption{\textbf{Performance on Longest6 benchmark.}
}
\label{tab:longest6}
\end{table}

\subsection{Comparsion with SOTA}
\vspace{-4pt}
We compare our work with state-of-the-art works in two competitive benchmarks of closed-loop evaluation. 
The results are shown in Tab.~\ref{tab:town05-long} and Tab.~\ref{tab:longest6}. 
We could observe that our model outperforms previous SOTA by a large margin on both benchmarks. Specifically, in Town5Long as shown in Tab.~\ref{tab:town05-long}, ThinkTwice achieves the best DS under both protocols while Roach and MILE could run for a long time (highest RC) but have much more collision or violation of traffic rules. On other hand, Transfuser runs most safely (highest IS) but is too cautious to complete the route. 
As for the Longest6 benchmark as shown in Tab.~\ref{tab:longest6}, which is proposed by Transfuser, they could 
obtain a very high route completion score but our method achieves the best DS and IS which suggests a much safer driving process.

\subsection{Component Analysis}
\vspace{-4pt}
In this section, we provide an empirical analysis of design choices among ThinkTwice. We use \textbf{Town05-Long} benchmark with 3 repeats to reduce the variance and we use 189K data of 4 towns, which means our model has never seen town05 during training.

\begin{table}[!h]
\small
\scalebox{0.95}{
\begin{tabular}{llccc}
\toprule
ID& Method                               & DS$\uparrow$          & RC$\uparrow$           & IS$\uparrow$           \\ \midrule
1 & Baseline                             & 43.0$\pm$3.9 & 74.2$\pm$3.2 & 0.56$\pm$0.05 \\ 
2 & \makecell[l]{+Segmentation \& \\Depth Task}                     & 45.0$\pm$2.2       & 73.2$\pm$4.0          & 0.60$\pm$0.04         \\ 
3 & \makecell[l]{+Segmentation \& \\Depth Projection}              & 57.8$\pm$1.5        & 78.8$\pm$3.3        & \textbf{0.74}$\pm$0.03         \\ 
4 & +Two Frames                          & \textbf{59.3}$\pm$1.4       & \textbf{80.2}$\pm$3.6         & \textbf{0.74}$\pm$0.04         \\ 
5 & \makecell[l]{Expert Feature\, -\textgreater{}\\BEV Segmentation} & 50.5$\pm$2.3        & 73.4$\pm$3.2        & 0.70$\pm$0.03         \\ \bottomrule
\end{tabular}}
\vspace{-5pt}
\caption{\textbf{Ablation for Encoder Design.} Baseline means original LSS\cite{philion2020lift} for camera inputs and SECOND~\cite{yan2018second} for LiDAR input with a TCP~\cite{wu2022trajectoryguided} head. Model2 adds the depth and segmentation auxiliary for images. Model3 additionally uses depth and segmentation during the projection to BEV. Model4 uses 2 frames as input. Model5 changes the BEV supervision from  BEV features of the expert to BEV segmentation task.
}
\label{tab:encoder}
\end{table}


\smallskip
\noindent\textbf{Encoder Design}: In Tab.~\ref{tab:encoder}, we give the performance of different encoder design choices in ThinkTwice.
From \textbf{Model1}, we can find that simply adopting geometric fusion without any relevant supervision leads to poor performance, which aligns with the conclusion in~\cite{Chitta2022PAMI}. 
\textbf{Model2} with depth and semantic segmentation tasks has slightly better results, which may come from the regularization effects of the two auxiliary tasks on the image features.
In \textbf{Model3}, the explicit usage of depth and segmentation prediction during the projection process from image features to BEV features boosts the performance significantly, which demonstrates the importance of supervised geometric projection. 
In \textbf{Model4}, we further use two frames as inputs instead of one. Contrary to intuition, the improvement is very marginal considering the motion clue is introduced. It might be related to the inertia/copycat problem~\cite{wen2020fighting,Chitta2022PAMI} where the model learns to cheat by simply copying the movement between the previous and current frame. It could lead to degenerated performance during closed-loop evaluation. However, since the \emph{Prediction Module} in the decoder requires the motion clues of surrounding agents, we keep the input as 2 frames in all the following experiments.
In \textbf{Model5}, we replace the Expert Feature Distillation to the BEV feature with the BEV segmentation task as in~\cite{chen2022lav,Chitta2022PAMI,hu2022model}. We can observe a performance drop which might be due to the fact that the BEV segmentation task could only serve as an implicit regularization while the expert feature contains decision-related information.

Due to Model4’s superior performance, we adopt it  as the encoder of ThinkTwice. 

\begin{table}[t!]
\small
\centering
\begin{tabular}{llccc}
\toprule
ID& Method                               & DS$\uparrow$          & RC$\uparrow$           & IS$\uparrow$           \\ \midrule
4 & TCP-Head    &                  59.3$\pm$1.4       & 80.2$\pm$3.6         & \textbf{0.74}$\pm$0.04 \\ 
5 & +1 Decoder                     & 61.6$\pm$1.4       & 90.8$\pm$2.2          & $ 0.68\pm$0.04         \\ 
6 & +5 Decoders              & \textbf{65.0}$\pm$1.7        & 95.5$\pm$2.0        & 0.69$\pm$0.05         \\ 
7 & w/o Look                          & 62.4$\pm$1.8       & 93.5$\pm$2.1         & 0.67$\pm$0.08         \\ 
8 & w/o Predict & 61.7$\pm$1.9        & 96.2$\pm$3.0        & 0.63$\pm$0.03         \\ 
9 & w/o TF & 62.0$\pm$2.2        & \textbf{97.1}$\pm$2.9        & 0.62$\pm$0.04         \\ \bottomrule
\end{tabular}
\vspace{-5pt}
\caption{\textbf{Ablation for Decoder Design.} The baseline is Model4 in the encoder. Model5 and Model6 add 1 and 5 proposed decoder modules respectively. Model7 removes the Look Module from Model6 while Model8 removes the Prediction Module from Model6. Model9 removes the teacher forcing technique. 
}
\label{tab:decoder}
\end{table}

\smallskip
\noindent\textbf{Decoder Design:}
Based on \textbf{Model4} in Tab.~\ref{tab:encoder}, we conduct component analysis for the decoder and the results are in Tab.~\ref{tab:decoder}. In \textbf{Model5}, we could observe performance improvement with one additional decoder layer. Specifically, RC improves a lot which indicates less stuck while IS drops a little bit which is natural since it has a larger possibility to have any events during the much longer driving process.  In \textbf{Model6}, with 5 stacked decoder layers, the results are significantly boosted, which demonstrates the effectiveness of the proposed decoder paradigm and its strong scalability. Until now, we obtain ThinkTwice's final model - Model6 and we further conduct ablation studies with it. In \textbf{Model7}, the removal of Look Modules causes an explicit performance drop. In \textbf{Model8}, without the Prediction Module, it has a much lower DS, slightly higher RC, and much lower IS, which indicates a more reckless agent who tends to simply drive forward and ignore the environment. It makes sense since it lacks the ability to know the ramification of its decision with the Prediction Module. In \textbf{Model9}, we verify the effectiveness of the teacher-forcing technique. Without it, the model exhibits similar behaviors with Model8, \ie, removing the prediction module. It is in line with expectations since there is no extra information injected if we use the Prediction Module without any supervision, which has no significant difference with an enlarged MLP.

In conclusion, we verify the claims and approaches proposed in Sec.~\ref{sec:approach}. We found that it is essential to add prior knowledge to the encoder and decoder. We also demonstrate the effectiveness of stacking decoder layers with dense supervision, which leads to our state-of-the-art performance.

\subsection{Discussion about Enlarging Model Capacity}
\vspace{-4pt}
In this section, we aim to investigate different ways of enlarging the capacity of an end-to-end autonomous driving model and compare them in a fair setting.

\smallskip
\noindent\textbf{Enlarge Encoder}: One natural idea is to enlarge the size of the encoder, which works perfectly well and demonstrates strong scalability in both natural language processing~\cite{Devlin2019BERTPO} and computer vision~\cite{he2016deep} fields.
However, it does not apply in the end-to-end autonomous model, which has been observed in the community.
%
In our early exploration experiments, we have conducted experiments with TCP~\cite{wu2022trajectoryguided}, a most recent SOTA method with single image input and it uses a single ResNet-34 as the encoder. It provides a single-variable environment to observe the effects of enlarging the encoder size.
The results are in Tab.~\ref{tab:tcp-encoder}. We could find that enlarging the original TCP from ResNet-34 to ResNet-101 causes a significant performance drop.
We conjecture the reason why simply enlarging the encoder does not work is that in end-to-end autonomous driving, the encoder is only responsible for processing the multi-sensor input while the decoder is responsible for finding out decision-related information. Large encoders could lead to better scene representation feature but it does not contribute much to the decision process. Actually, it is the major motivation of ThinkTwice: enlarging the capacity of the decoder in a proper way.
\newline

\begin{table}[t!]
\small
\centering
\begin{tabular}{lccc}
\toprule
Encoder                               & DS$\uparrow$          & RC$\uparrow$           & IS$\uparrow$           \\ \midrule
ResNet-18   &                  48.7$\pm$1.7       & \textbf{81.3}$\pm$2.2         & 0.58$\pm$0.04 \\ 
ResNet-34   &                  \textbf{57.2}$\pm$1.5       & 80.4$\pm$1.5         & \textbf{0.73}$\pm$0.02 \\ 
ResNet-101   &                  38.0$\pm$4.5       & 79.9$\pm$3.7         & 0.49$\pm$0.02 \\ \bottomrule
\end{tabular}
\vspace{-5pt}
\caption{\textbf{Performance of  TCP~\cite{wu2022trajectoryguided} under different encoder sizes.}
}
\label{tab:tcp-encoder}
\end{table}

\begin{table}[t!]
\small
\centering
\begin{tabular}{lcccc}
\toprule
Method & \#Param                              & DS$\uparrow$          & RC$\uparrow$           & IS$\uparrow$           \\ \midrule
Baseline   &  78.2M & 59.3$\pm$1.4       & 80.2$\pm$3.6   &  \textbf{0.74}$\pm$0.04 \\ \midrule
Ours   & 120.2M  &          \textbf{65.0}$\pm$1.7        & \textbf{95.5}$\pm$2.0        & 0.69$\pm$0.05 \\ 
MLP/GRU   & 123.3M &                59.4$\pm$2.4       & 81.2$\pm$3.0         & 0.72$\pm$0.04 \\ 
Backbone   &    119.3M &                    58.4$\pm$3.1       & 80.0$\pm$3.6 & 0.73$\pm$0.06  \\ \bottomrule
\end{tabular}
\vspace{-5pt}
\caption{\textbf{Performance of ThinkTwice with different major parts under the same number of parameters.} \emph{Baseline} uses the Model4 as the encoder and TCP head as the decoder. \emph{Ours} uses 5 stacked proposed decoder layers with supervision. \emph{MLP/GRU} uses the TCP head with  2x depth and 4x width. \emph{Backbone} uses ResNet-152 as the backbone instead of ResNet-50.}
 \label{tab:large}
\end{table}

\smallskip
\noindent\textbf{Enlarge MLP/GRU}: Besides enlarging the encoder size and the proposed stacking decoders, another choice to increase the models' capacity is to increase the width/depth of the classical MLP/GRU decoder. Here, we take Model4 as the baseline which includes the proposed encoder and a TCP head and we enlarge the model in the three aforementioned ways. The results are in Tab.~\ref{tab:large}. We can observe that simply increasing the depth/width of the encoder or the decoder would not bring performance gain. On the contrary,  ThinkTwice enlarges the decoder's capacity in a coarse-to-fine fashion with dense supervision and spatial-temporal knowledge, which injects strong priors into the model and thus leads to better performance.

\section{Conclusion}
\vspace{-4pt}
In this work, we present ThinkTwice, an end-to-end autonomous driving paradigm that emphasizes enlarging the capacity of the decoder. We propose a scalable decoder layer with dense supervision and spatial-temporal priors. By stacking the proposed decoder layer, we achieve state-of-the-art performance on two competitive closed-loop autonomous driving benchmarks.
We hope the attempts and successful parts illustrated in this study could provide useful information for this line of study in the community.


\smallskip
\noindent\textbf{Acknowledgements.} 
We are grateful to Mingjie Pan for technical assistance. We also thank the reviewers for their constructive comments and suggestions.
%
This work was in part supported by NSFC (62206172, 62222607), Shanghai Municipal Science and Technology Major Project (2021SHZDZX0102), and Shanghai Committee of Science and Technology (21DZ1100100).

\clearpage
\newpage
{\small
\bibliographystyle{ieee_fullname}
\bibliography{egbib}

\begin{thebibliography}{10}\itemsep=-1pt

\bibitem{bansal2018chauffeurnet}
Mayank Bansal, Alex Krizhevsky, and Abhijit Ogale.
\newblock Chauffeurnet: Learning to drive by imitating the best and
  synthesizing the worst.
\newblock {\em arXiv preprint arXiv:1812.03079}, 2018.

\bibitem{bengio2015scheduled}
Samy Bengio, Oriol Vinyals, Navdeep Jaitly, and Noam Shazeer.
\newblock Scheduled sampling for sequence prediction with recurrent neural
  networks.
\newblock {\em Advances in neural information processing systems}, 28, 2015.

\bibitem{nvidia2016EndTE}
Mariusz Bojarski, Davide~Del Testa, Daniel Dworakowski, Bernhard Firner, Beat
  Flepp, Prasoon Goyal, Lawrence~D. Jackel, Mathew Monfort, Urs~A. Muller,
  Jiakai Zhang, Xin Zhang, Jake Zhao, and Karol Zieba.
\newblock End to end learning for self-driving cars.
\newblock {\em arXiv preprint arXiv:1604.07316}, 2016.

\bibitem{can2021structured}
Yigit~Baran Can, Alexander Liniger, Danda~Pani Paudel, and Luc Van~Gool.
\newblock Structured bird's-eye-view traffic scene understanding from onboard
  images.
\newblock In {\em Proceedings of the IEEE/CVF International Conference on
  Computer Vision}, pages 15661--15670, 2021.

\bibitem{carion2020end}
Nicolas Carion, Francisco Massa, Gabriel Synnaeve, Nicolas Usunier, Alexander
  Kirillov, and Sergey Zagoruyko.
\newblock End-to-end object detection with transformers.
\newblock In {\em European conference on computer vision}, pages 213--229.
  Springer, 2020.

\bibitem{casas2021mp3}
Sergio Casas, Abbas Sadat, and Raquel Urtasun.
\newblock Mp3: A unified model to map, perceive, predict and plan.
\newblock In {\em Proceedings of the IEEE/CVF Conference on Computer Vision and
  Pattern Recognition}, pages 14403--14412, 2021.

\bibitem{chekroun2021gri}
Raphael Chekroun, Marin Toromanoff, Sascha Hornauer, and Fabien Moutarde.
\newblock Gri: General reinforced imitation and its application to vision-based
  autonomous driving.
\newblock {\em arXiv preprint arXiv:2111.08575}, 2021.

\bibitem{chen2021learning}
Dian Chen, Vladlen Koltun, and Philipp Kr{\"a}henb{\"u}hl.
\newblock Learning to drive from a world on rails.
\newblock In {\em Proceedings of the IEEE/CVF International Conference on
  Computer Vision}, 2021.

\bibitem{chen2022lav}
Dian Chen and Philipp Kr{\"a}henb{\"u}hl.
\newblock Learning from all vehicles.
\newblock In {\em Proceedings of the IEEE Conference on Computer Vision and
  Pattern Recognition}, 2022.

\bibitem{chen2020learning}
Dian Chen, Brady Zhou, Vladlen Koltun, and Philipp Kr{\"a}henb{\"u}hl.
\newblock Learning by cheating.
\newblock In {\em Conference on Robot Learning}, pages 66--75. PMLR, 2020.

\bibitem{chen2022persformer}
Li Chen, Chonghao Sima, Yang Li, Zehan Zheng, Jiajie Xu, Xiangwei Geng,
  Hongyang Li, Conghui He, Jianping Shi, Yu Qiao, et~al.
\newblock Persformer: 3d lane detection via perspective transformer and the
  openlane benchmark.
\newblock In {\em European Conference on Computer Vision}, pages 550--567.
  Springer, 2022.

\bibitem{chitta2021neat}
Kashyap Chitta, Aditya Prakash, and Andreas Geiger.
\newblock Neat: Neural attention fields for end-to-end autonomous driving.
\newblock In {\em Proceedings of the IEEE/CVF International Conference on
  Computer Vision}, pages 15793--15803, 2021.

\bibitem{Chitta2022PAMI}
Kashyap Chitta, Aditya Prakash, Bernhard Jaeger, Zehao Yu, Katrin Renz, and
  Andreas Geiger.
\newblock Transfuser: Imitation with transformer-based sensor fusion for
  autonomous driving.
\newblock {\em Pattern Analysis and Machine Intelligence (PAMI)}, 2022.

\bibitem{cho2014properties}
Kyunghyun Cho, Bart van Merri{\"e}nboer, Dzmitry Bahdanau, and Yoshua Bengio.
\newblock On the properties of neural machine translation: Encoder--decoder
  approaches.
\newblock In {\em Proceedings of SSST-8, Eighth Workshop on Syntax, Semantics
  and Structure in Statistical Translation}, pages 103--111, 2014.

\bibitem{codevilla2018cil}
Felipe Codevilla, Matthias M{\"u}ller, Antonio L{\'o}pez, Vladlen Koltun, and
  Alexey Dosovitskiy.
\newblock End-to-end driving via conditional imitation learning.
\newblock In {\em 2018 IEEE international conference on robotics and
  automation}, pages 4693--4700. IEEE, 2018.

\bibitem{codevilla2019exploring}
Felipe Codevilla, Eder Santana, Antonio~M L{\'o}pez, and Adrien Gaidon.
\newblock Exploring the limitations of behavior cloning for autonomous driving.
\newblock In {\em Proceedings of the IEEE/CVF International Conference on
  Computer Vision}, pages 9329--9338, 2019.

\bibitem{10.1093/mnras/79.5.384}
A.~E. Conrady.
\newblock {Decentred Lens-Systems}.
\newblock {\em Monthly Notices of the Royal Astronomical Society},
  79(5):384--390, 03 1919.

\bibitem{mmcv}
MMCV Contributors.
\newblock {MMCV: OpenMMLab} computer vision foundation.
\newblock \url{https://github.com/open-mmlab/mmcv}, 2018.

\bibitem{cui2021lookout}
Alexander Cui, Sergio Casas, Abbas Sadat, Renjie Liao, and Raquel Urtasun.
\newblock Lookout: Diverse multi-future prediction and planning for
  self-driving.
\newblock In {\em Proceedings of the IEEE/CVF International Conference on
  Computer Vision}, pages 16107--16116, 2021.

\bibitem{deng2021voxel}
Jiajun Deng, Shaoshuai Shi, Peiwei Li, Wengang Zhou, Yanyong Zhang, and
  Houqiang Li.
\newblock Voxel r-cnn: Towards high performance voxel-based 3d object
  detection.
\newblock In {\em Proceedings of the AAAI Conference on Artificial
  Intelligence}, 2021.

\bibitem{deng18r}
Zijun Deng, Xiaowei Hu, Lei Zhu, Xuemiao Xu, Jing Qin, Guoqiang Han, and
  Pheng-Ann Heng.
\newblock R$^{3}${N}et: Recurrent residual refinement network for saliency
  detection.
\newblock In {\em IJCAI}, 2018.

\bibitem{Devlin2019BERTPO}
Jacob Devlin, Ming-Wei Chang, Kenton Lee, and Kristina Toutanova.
\newblock Bert: Pre-training of deep bidirectional transformers for language
  understanding.
\newblock {\em arXiv preprint arXiv:1810.04805}, 2018.

\bibitem{Dosovitskiy17}
Alexey Dosovitskiy, German Ros, Felipe Codevilla, Antonio Lopez, and Vladlen
  Koltun.
\newblock {CARLA}: {An} open urban driving simulator.
\newblock In {\em Proceedings of the 1st Annual Conference on Robot Learning},
  pages 1--16, 2017.

\bibitem{fischer2015flownet}
Philipp Fischer, Alexey Dosovitskiy, Eddy Ilg, Philip H{\"a}usser, Caner
  Haz{\i}rba{\c{s}}, Vladimir Golkov, Patrick Van~der Smagt, Daniel Cremers,
  and Thomas Brox.
\newblock Flownet: Learning optical flow with convolutional networks.
\newblock {\em arXiv preprint arXiv:1504.06852}, 2015.

\bibitem{garnett20193d}
Noa Garnett, Rafi Cohen, Tomer Pe'er, Roee Lahav, and Dan Levi.
\newblock 3d-lanenet: end-to-end 3d multiple lane detection.
\newblock In {\em Proceedings of the IEEE/CVF International Conference on
  Computer Vision}, pages 2921--2930, 2019.

\bibitem{girshick2015fast}
Ross Girshick.
\newblock Fast r-cnn.
\newblock In {\em Proceedings of the IEEE International Conference on Computer
  Vision}, pages 1440--1448, 2015.

\bibitem{girshick14CVPRRCNN}
Ross Girshick, Jeff Donahue, Trevor Darrell, and Jitendra Malik.
\newblock Rich feature hierarchies for accurate object detection and semantic
  segmentation.
\newblock In {\em Proceedings of the IEEE/CVF Conference on Computer Vision and
  Pattern Recognition}, 2014.

\bibitem{he2016deep}
Kaiming He, Xiangyu Zhang, Shaoqing Ren, and Jian Sun.
\newblock Deep residual learning for image recognition.
\newblock In {\em Proceedings of the IEEE Conference on Computer Vision and
  Pattern Recognition}, pages 770--778, 2016.

\bibitem{hu2022model}
Anthony Hu, Gianluca Corrado, Nicolas Griffiths, Zak Murez, Corina Gurau,
  Hudson Yeo, Alex Kendall, Roberto Cipolla, and Jamie Shotton.
\newblock Model-based imitation learning for urban driving.
\newblock {\em arXiv preprint arXiv:2210.07729}, 2022.

\bibitem{hu2022st}
Shengchao Hu, Li Chen, Penghao Wu, Hongyang Li, Junchi Yan, and Dacheng Tao.
\newblock St-p3: End-to-end vision-based autonomous driving via
  spatial-temporal feature learning.
\newblock In {\em European Conference on Computer Vision}, pages 533--549.
  Springer, 2022.

\bibitem{Hu2022AFDetV2RT}
Yihan Hu, Zhuangzhuang Ding, Runzhou Ge, Wenxin Shao, Li Huang, Kun Li, and
  Qiang Liu.
\newblock Afdetv2: Rethinking the necessity of the second stage for object
  detection from point clouds.
\newblock In {\em Proceedings of the AAAI Conference on Artificial
  Intelligence}, 2022.

\bibitem{uniad}
Yihan Hu, Jiazhi Yang, Li Chen, Keyu Li, Chonghao Sima, Xizhou Zhu, Siqi Chai,
  Senyao Du, Tianwei Lin, Wenhai Wang, Lewei Lu, Xiaosong Jia, Qiang Liu,
  Jifeng Dai, Yu Qiao, and Hongyang Li.
\newblock Planning-oriented autonomous driving.
\newblock In {\em Proceedings of the IEEE/CVF Conference on Computer Vision and
  Pattern Recognition}, 2022.

\bibitem{huang2022bevpoolv2}
Junjie Huang and Guan Huang.
\newblock Bevpoolv2: A cutting-edge implementation of bevdet toward deployment.
\newblock {\em arXiv preprint arXiv:2211.17111}, 2022.

\bibitem{huang2021bevdet}
Junjie Huang, Guan Huang, Zheng Zhu, and Dalong Du.
\newblock Bevdet: High-performance multi-camera 3d object detection in
  bird-eye-view.
\newblock {\em arXiv preprint arXiv:2112.11790}, 2021.

\bibitem{ilg2017flownet}
Eddy Ilg, Nikolaus Mayer, Tonmoy Saikia, Margret Keuper, Alexey Dosovitskiy,
  and Thomas Brox.
\newblock Flownet 2.0: Evolution of optical flow estimation with deep networks.
\newblock In {\em Proceedings of the IEEE Conference on Computer Vision and
  Pattern Recognition}, pages 2462--2470, 2017.

\bibitem{jia2022towards}
Xiaosong Jia, Li Chen, Penghao Wu, Jia Zeng, Junchi Yan, Hongyang Li, and Yu
  Qiao.
\newblock Towards capturing the temporal dynamics for trajectory prediction: a
  coarse-to-fine approach.
\newblock In {\em 6th Annual Conference on Robot Learning}, 2022.

\bibitem{jia2021ide}
Xiaosong Jia, Liting Sun, Masayoshi Tomizuka, and Wei Zhan.
\newblock Ide-net: Interactive driving event and pattern extraction from human
  data.
\newblock {\em IEEE Robotics and Automation Letters}, 6(2):3065--3072, 2021.

\bibitem{pmlr-v164-jia22a}
Xiaosong Jia, Liting Sun, Hang Zhao, Masayoshi Tomizuka, and Wei Zhan.
\newblock Multi-agent trajectory prediction by combining egocentric and
  allocentric views.
\newblock In Aleksandra Faust, David Hsu, and Gerhard Neumann, editors, {\em
  Proceedings of the 5th Conference on Robot Learning}, volume 164 of {\em
  Proceedings of Machine Learning Research}, pages 1434--1443. PMLR, 08--11 Nov
  2022.

\bibitem{li2022delving}
Hongyang Li, Chonghao Sima, Jifeng Dai, Wenhai Wang, Lewei Lu, Huijie Wang,
  Enze Xie, Zhiqi Li, Hanming Deng, Hao Tian, et~al.
\newblock Delving into the devils of bird's-eye-view perception: A review,
  evaluation and recipe.
\newblock {\em arXiv preprint arXiv:2209.05324}, 2022.

\bibitem{li2022bevstereo}
Yinhao Li, Han Bao, Zheng Ge, Jinrong Yang, Jianjian Sun, and Zeming Li.
\newblock Bevstereo: Enhancing depth estimation in multi-view 3d object
  detection with dynamic temporal stereo.
\newblock {\em arXiv preprint arXiv:2209.10248}, 2022.

\bibitem{li2022bevdepth}
Yinhao Li, Zheng Ge, Guanyi Yu, Jinrong Yang, Zengran Wang, Yukang Shi,
  Jianjian Sun, and Zeming Li.
\newblock Bevdepth: Acquisition of reliable depth for multi-view 3d object
  detection.
\newblock {\em arXiv preprint arXiv:2206.10092}, 2022.

\bibitem{li2021lidar}
Zhichao Li, Feng Wang, and Naiyan Wang.
\newblock Lidar r-cnn: An efficient and universal 3d object detector.
\newblock In {\em Proceedings of the IEEE/CVF Conference on Computer Vision and
  Pattern Recognition}, pages 7546--7555, 2021.

\bibitem{li2022bevformer}
Zhiqi Li, Wenhai Wang, Hongyang Li, Enze Xie, Chonghao Sima, Tong Lu, Qiao Yu,
  and Jifeng Dai.
\newblock Bevformer: Learning bird's-eye-view representation from multi-camera
  images via spatiotemporal transformers.
\newblock In {\em European Conference on Computer Vision}, 2022.

\bibitem{7780449}
N. Liu and J. Han.
\newblock Dhsnet: Deep hierarchical saliency network for salient object
  detection.
\newblock In {\em Proceedings of the IEEE Conference on Computer Vision and
  Pattern Recognition}, 2016.

\bibitem{liu2018path}
Shu Liu, Lu Qi, Haifang Qin, Jianping Shi, and Jiaya Jia.
\newblock Path aggregation network for instance segmentation.
\newblock In {\em Proceedings of the IEEE conference on computer vision and
  pattern recognition}, pages 8759--8768, 2018.

\bibitem{liu2022petr}
Yingfei Liu, Tiancai Wang, Xiangyu Zhang, and Jian Sun.
\newblock Petr: Position embedding transformation for multi-view 3d object
  detection.
\newblock {\em arXiv preprint arXiv:2203.05625}, 2022.

\bibitem{liu2022convnet}
Zhuang Liu, Hanzi Mao, Chao-Yuan Wu, Christoph Feichtenhofer, Trevor Darrell,
  and Saining Xie.
\newblock A convnet for the 2020s.
\newblock {\em Proceedings of the IEEE/CVF Conference on Computer Vision and
  Pattern Recognition (CVPR)}, 2022.

\bibitem{liu2022bevfusion}
Zhijian Liu, Haotian Tang, Alexander Amini, Xinyu Yang, Huizi Mao, Daniela Rus,
  and Song Han.
\newblock Bevfusion: Multi-task multi-sensor fusion with unified bird's-eye
  view representation.
\newblock {\em arXiv preprint arXiv:2205.13542}, 2022.

\bibitem{loshchilov2018decoupled}
Ilya Loshchilov and Frank Hutter.
\newblock Decoupled weight decay regularization.
\newblock In {\em International Conference on Learning Representations}, 2018.

\bibitem{malla2020titan}
Srikanth Malla, Behzad Dariush, and Chiho Choi.
\newblock Titan: Future forecast using action priors.
\newblock In {\em Proceedings of the IEEE/CVF Conference on Computer Vision and
  Pattern Recognition}, pages 11186--11196, 2020.

\bibitem{mangalam2021goals}
Karttikeya Mangalam, Yang An, Harshayu Girase, and Jitendra Malik.
\newblock From goals, waypoints \& paths to long term human trajectory
  forecasting.
\newblock In {\em Proceedings of the IEEE/CVF International Conference on
  Computer Vision}, pages 15233--15242, 2021.

\bibitem{marchetti2020mantra}
Francesco Marchetti, Federico Becattini, Lorenzo Seidenari, and Alberto~Del
  Bimbo.
\newblock Mantra: Memory augmented networks for multiple trajectory prediction.
\newblock In {\em Proceedings of the IEEE/CVF Conference on Computer Vision and
  Pattern Recognition}, pages 7143--7152, 2020.

\bibitem{muller2005off}
Urs Muller, Jan Ben, Eric Cosatto, Beat Flepp, and Yann Cun.
\newblock Off-road obstacle avoidance through end-to-end learning.
\newblock {\em Advances in neural information processing systems}, 18, 2005.

\bibitem{paszke2019pytorch}
Adam Paszke, Sam Gross, Francisco Massa, Adam Lerer, James Bradbury, Gregory
  Chanan, Trevor Killeen, Zeming Lin, Natalia Gimelshein, Luca Antiga, et~al.
\newblock Pytorch: An imperative style, high-performance deep learning library.
\newblock {\em Advances in neural information processing systems}, 32, 2019.

\bibitem{philion2020lift}
Jonah Philion and Sanja Fidler.
\newblock Lift, splat, shoot: Encoding images from arbitrary camera rigs by
  implicitly unprojecting to 3d.
\newblock In {\em European Conference on Computer Vision}, pages 194--210.
  Springer, 2020.

\bibitem{pomerleau1988alvinn}
Dean~A Pomerleau.
\newblock Alvinn: An autonomous land vehicle in a neural network.
\newblock {\em Advances in neural information processing systems}, 1, 1988.

\bibitem{Prakash2021CVPR}
Aditya Prakash, Kashyap Chitta, and Andreas Geiger.
\newblock Multi-modal fusion transformer for end-to-end autonomous driving.
\newblock In {\em Proceedings of the IEEE Conference on Computer Vision and
  Pattern Recognition}, 2021.

\bibitem{Qin_2019_CVPR}
Xuebin Qin, Zichen Zhang, Chenyang Huang, Chao Gao, Masood Dehghan, and Martin
  Jagersand.
\newblock Basnet: Boundary-aware salient object detection.
\newblock In {\em Proceedings of the IEEE Conference on Computer Vision and
  Pattern Recognition}, 2019.

\bibitem{reading2021categorical}
Cody Reading, Ali Harakeh, Julia Chae, and Steven~L Waslander.
\newblock Categorical depth distribution network for monocular 3d object
  detection.
\newblock In {\em Proceedings of the IEEE/CVF Conference on Computer Vision and
  Pattern Recognition}, pages 8555--8564, 2021.

\bibitem{renNIPS15fasterrcnn}
Shaoqing Ren, Kaiming He, Ross Girshick, and Jian Sun.
\newblock Faster {R-CNN}: Towards real-time object detection with region
  proposal networks.
\newblock In {\em Advances in Neural Information Processing Systems}, 2015.

\bibitem{Renz2022CORL}
Katrin Renz, Kashyap Chitta, Otniel-Bogdan Mercea, A.~Sophia Koepke, Zeynep
  Akata, and Andreas Geiger.
\newblock Plant: Explainable planning transformers via object-level
  representations.
\newblock In {\em Conference on Robotic Learning}, 2022.

\bibitem{roddick2018orthographic}
Thomas Roddick, Alex Kendall, and Roberto Cipolla.
\newblock Orthographic feature transform for monocular 3d object detection.
\newblock {\em arXiv preprint arXiv:1811.08188}, 2018.

\bibitem{ronneberger2015u}
Olaf Ronneberger, Philipp Fischer, and Thomas Brox.
\newblock U-net: Convolutional networks for biomedical image segmentation.
\newblock In {\em International Conference on Medical image computing and
  computer-assisted intervention}, pages 234--241. Springer, 2015.

\bibitem{sadat2020perceive}
Abbas Sadat, Sergio Casas, Mengye Ren, Xinyu Wu, Pranaab Dhawan, and Raquel
  Urtasun.
\newblock Perceive, predict, and plan: Safe motion planning through
  interpretable semantic representations.
\newblock In {\em European Conference on Computer Vision}, pages 414--430.
  Springer, 2020.

\bibitem{sauer2018cal}
Axel Sauer, Nikolay Savinov, and Andreas Geiger.
\newblock Conditional affordance learning for driving in urban environments.
\newblock In {\em Conference on Robot Learning}, pages 237--252. PMLR, 2018.

\bibitem{shao2022interfuser}
Hao Shao, Letian Wang, RuoBing Chen, Hongsheng Li, and Yu Liu.
\newblock Safety-enhanced autonomous driving using interpretable sensor fusion
  transformer.
\newblock {\em arXiv preprint arXiv:2207.14024}, 2022.

\bibitem{shi2020pv}
Shaoshuai Shi, Chaoxu Guo, Li Jiang, Zhe Wang, Jianping Shi, Xiaogang Wang, and
  Hongsheng Li.
\newblock Pv-rcnn: Point-voxel feature set abstraction for 3d object detection.
\newblock In {\em Proceedings of the IEEE/CVF Conference on Computer Vision and
  Pattern Recognition}, pages 10529--10538, 2020.

\bibitem{shi2022motion}
Shaoshuai Shi, Li Jiang, Dengxin Dai, and Bernt Schiele.
\newblock Motion transformer with global intention localization and local
  movement refinement.
\newblock {\em arXiv preprint arXiv:2209.13508}, 2022.

\bibitem{shi2019pointrcnn}
Shaoshuai Shi, Xiaogang Wang, and Hongsheng Li.
\newblock Pointrcnn: 3d object proposal generation and detection from point
  cloud.
\newblock In {\em Proceedings of the IEEE/CVF Conference on Computer Vision and
  Pattern Recognition}, pages 770--779, 2019.

\bibitem{IBISCape22}
Abanob Soliman, Fabien Bonardi, D{\'e}sir{\'e} Sidib{\'e}, and Samia Bouchafa.
\newblock {IBISCape}: A simulated benchmark for multi-modal {SLAM} systems
  evaluation in large-scale dynamic environments.
\newblock {\em Journal of Intelligent {\&} Robotic Systems}, 106(3):53, Oct
  2022.

\bibitem{sun2018pwc}
Deqing Sun, Xiaodong Yang, Ming-Yu Liu, and Jan Kautz.
\newblock Pwc-net: Cnns for optical flow using pyramid, warping, and cost
  volume.
\newblock In {\em Proceedings of the IEEE Conference on Computer Vision and
  Pattern Recognition}, pages 8934--8943, 2018.

\bibitem{sun2021rsn}
Pei Sun, Weiyue Wang, Yuning Chai, Gamaleldin Elsayed, Alex Bewley, Xiao Zhang,
  Cristian Sminchisescu, and Dragomir Anguelov.
\newblock Rsn: Range sparse net for efficient, accurate lidar 3d object
  detection.
\newblock In {\em Proceedings of the IEEE/CVF Conference on Computer Vision and
  Pattern Recognition}, pages 5725--5734, 2021.

\bibitem{teed2020raft}
Zachary Teed and Jia Deng.
\newblock Raft: Recurrent all-pairs field transforms for optical flow.
\newblock In {\em European Conference on Computer Vision}, pages 402--419.
  Springer, 2020.

\bibitem{toromanoff2020end}
Marin Toromanoff, Emilie Wirbel, and Fabien Moutarde.
\newblock End-to-end model-free reinforcement learning for urban driving using
  implicit affordances.
\newblock In {\em Proceedings of the IEEE/CVF Conference on Computer Vision and
  Pattern Recognition}, pages 7153--7162, 2020.

\bibitem{wangiccv17}
Tiantian Wang, Ali Borji, Lihe Zhang, Pingping Zhang, and Huchuan Lu.
\newblock A stagewise refinement model for detecting salient objects in images.
\newblock In {\em The IEEE International Conference on Computer Vision}, 2017.

\bibitem{wen2020fighting}
Chuan Wen, Jierui Lin, Trevor Darrell, Dinesh Jayaraman, and Yang Gao.
\newblock Fighting copycat agents in behavioral cloning from observation
  histories.
\newblock {\em Advances in Neural Information Processing Systems},
  33:2564--2575, 2020.

\bibitem{wu2023policy}
Penghao Wu, Li Chen, Hongyang Li, Xiaosong Jia, Junchi Yan, and Yu Qiao.
\newblock Policy pre-training for end-to-end autonomous driving via
  self-supervised geometric modeling.
\newblock {\em International Conference on Learning Representations}, 2023.

\bibitem{wu2022trajectoryguided}
Penghao Wu, Xiaosong Jia, Li Chen, Junchi Yan, Hongyang Li, and Yu Qiao.
\newblock Trajectory-guided control prediction for end-to-end autonomous
  driving: A simple yet strong baseline.
\newblock {\em Advances in Neural Information Processing Systems}, 2022.

\bibitem{xie2022m}
Enze Xie, Zhiding Yu, Daquan Zhou, Jonah Philion, Anima Anandkumar, Sanja
  Fidler, Ping Luo, and Jose~M Alvarez.
\newblock M\^{} 2bev: Multi-camera joint 3d detection and segmentation with
  unified birds-eye view representation.
\newblock {\em arXiv preprint arXiv:2204.05088}, 2022.

\bibitem{yan2018second}
Yan Yan, Yuxing Mao, and Bo Li.
\newblock Second: Sparsely embedded convolutional detection.
\newblock {\em Sensors}, 18(10):3337, 2018.

\bibitem{yang2019std}
Zetong Yang, Yanan Sun, Shu Liu, Xiaoyong Shen, and Jiaya Jia.
\newblock Std: Sparse-to-dense 3d object detector for point cloud.
\newblock In {\em Proceedings of the IEEE/CVF international conference on
  computer vision}, pages 1951--1960, 2019.

\bibitem{yin2021center}
Tianwei Yin, Xingyi Zhou, and Philipp Krahenbuhl.
\newblock Center-based 3d object detection and tracking.
\newblock In {\em Proceedings of the IEEE/CVF Conference on Computer Vision and
  Pattern Recognition}, pages 11784--11793, 2021.

\bibitem{zeng2019end}
Wenyuan Zeng, Wenjie Luo, Simon Suo, Abbas Sadat, Bin Yang, Sergio Casas, and
  Raquel Urtasun.
\newblock End-to-end interpretable neural motion planner.
\newblock In {\em Proceedings of the IEEE/CVF Conference on Computer Vision and
  Pattern Recognition}, pages 8660--8669, 2019.

\bibitem{zeng2020dsdnet}
Wenyuan Zeng, Shenlong Wang, Renjie Liao, Yun Chen, Bin Yang, and Raquel
  Urtasun.
\newblock Dsdnet: Deep structured self-driving network.
\newblock In {\em European Conference on Computer Vision}, pages 156--172.
  Springer, 2020.

\bibitem{zhang2022dino}
Hao Zhang, Feng Li, Shilong Liu, Lei Zhang, Hang Su, Jun Zhu, Lionel~M Ni, and
  Heung-Yeung Shum.
\newblock Dino: Detr with improved denoising anchor boxes for end-to-end object
  detection.
\newblock {\em arXiv preprint arXiv:2203.03605}, 2022.

\bibitem{zhang2022mmfn}
Qingwen Zhang, Mingkai Tang, Ruoyu Geng, Feiyi Chen, Ren Xin, and Lujia Wang.
\newblock Mmfn: Multi-modal-fusion-net for end-to-end driving.
\newblock {\em arXiv preprint arXiv:2207.00186}, 2022.

\bibitem{zhang2021roach}
Zhejun Zhang, Alexander Liniger, Dengxin Dai, Fisher Yu, and Luc Van~Gool.
\newblock End-to-end urban driving by imitating a reinforcement learning coach.
\newblock In {\em Proceedings of the IEEE/CVF International Conference on
  Computer Vision}, 2021.

\bibitem{zhou2022cross}
Brady Zhou and Philipp Kr{\"a}henb{\"u}hl.
\newblock Cross-view transformers for real-time map-view semantic segmentation.
\newblock In {\em Proceedings of the IEEE/CVF Conference on Computer Vision and
  Pattern Recognition}, pages 13760--13769, 2022.

\bibitem{zhou2018voxelnet}
Yin Zhou and Oncel Tuzel.
\newblock Voxelnet: End-to-end learning for point cloud based 3d object
  detection.
\newblock In {\em Proceedings of the IEEE Conference on Computer Vision and
  Pattern Recognition}, pages 4490--4499, 2018.

\bibitem{zhu2020deformable}
Xizhou Zhu, Weijie Su, Lewei Lu, Bin Li, Xiaogang Wang, and Jifeng Dai.
\newblock Deformable detr: Deformable transformers for end-to-end object
  detection.
\newblock {\em arXiv preprint arXiv:2010.04159}, 2020.

\end{thebibliography}
}

\clearpage
\newpage
\appendix
\noindent\textbf{\Large{Appendix}}

\begin{figure*}[!t]
		\centering
		\begin{subfigure}[t]{0.49\textwidth}
		\centering
		\includegraphics[width=8.5cm]{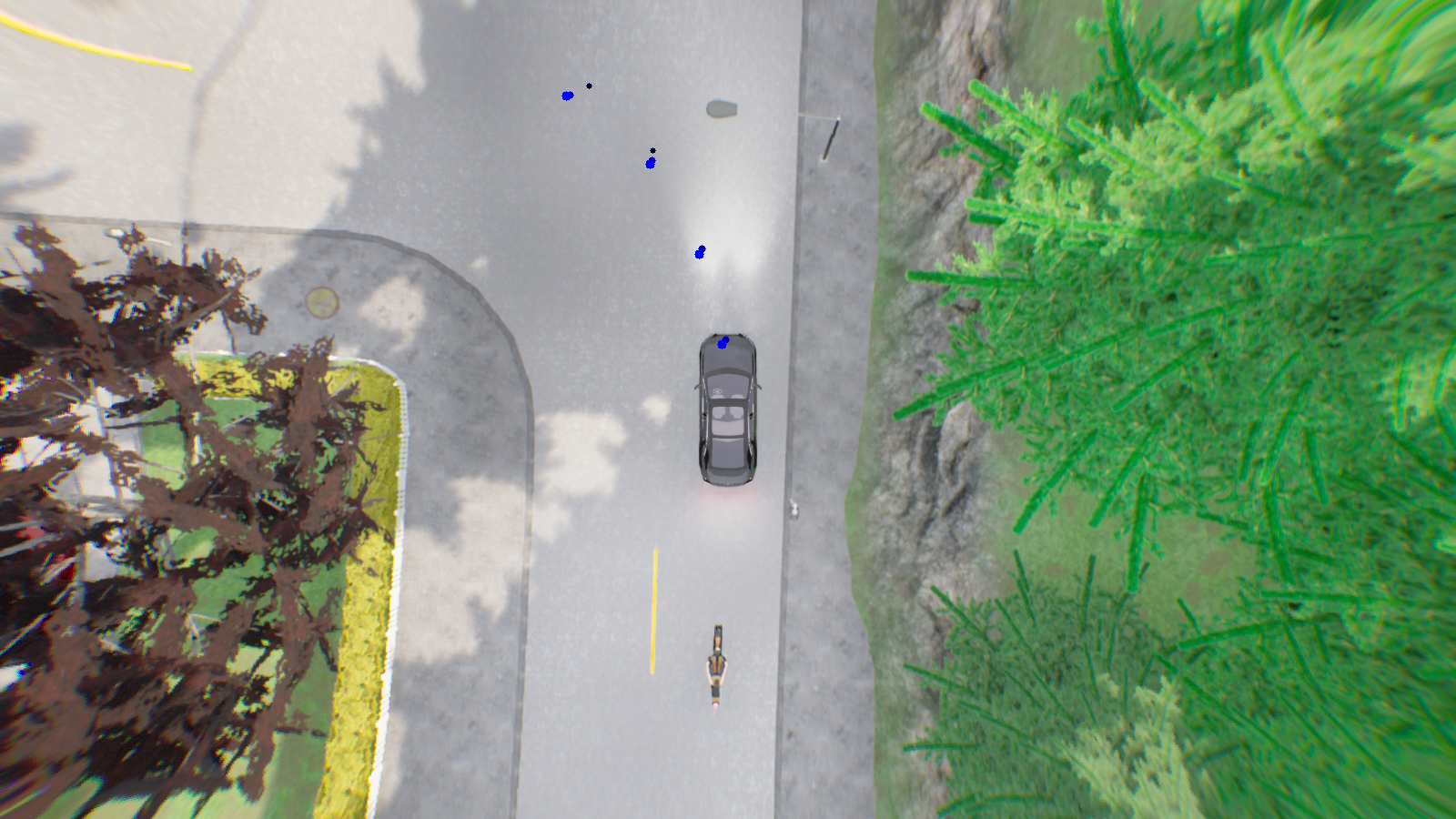}
		\captionsetup{width=8.5cm}
		\caption{\textbf{Quarter Turn}. It is a common failure case due to its sharp turn angle. Here, the refined trajectory has a more accurate turning route.}
		\end{subfigure}%
		\begin{subfigure}[t]{0.49\textwidth}
			\centering
			\includegraphics[width=8.5cm]{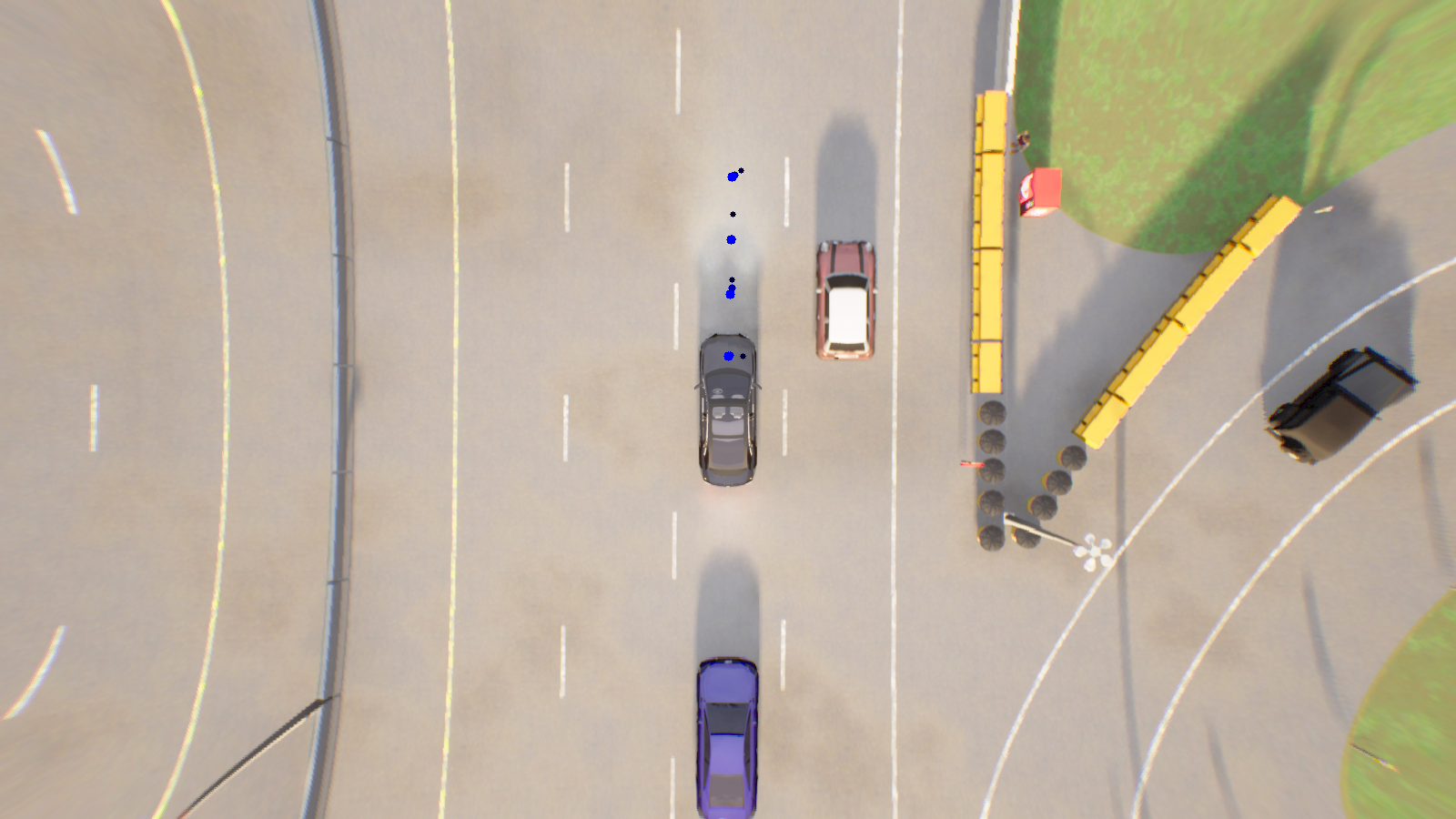}
			\captionsetup{width=8.5cm}
			\caption{\textbf{Emergency Stop}. For the jay-walker with the nearby vehicle's occlusion, the refined trajectory leads to a deacceleration compared to the original one.}
		\end{subfigure}%
		
		\begin{subfigure}[t]{0.49\textwidth}
		\centering
		\captionsetup{width=8.5cm}
		\includegraphics[width=8.5cm]{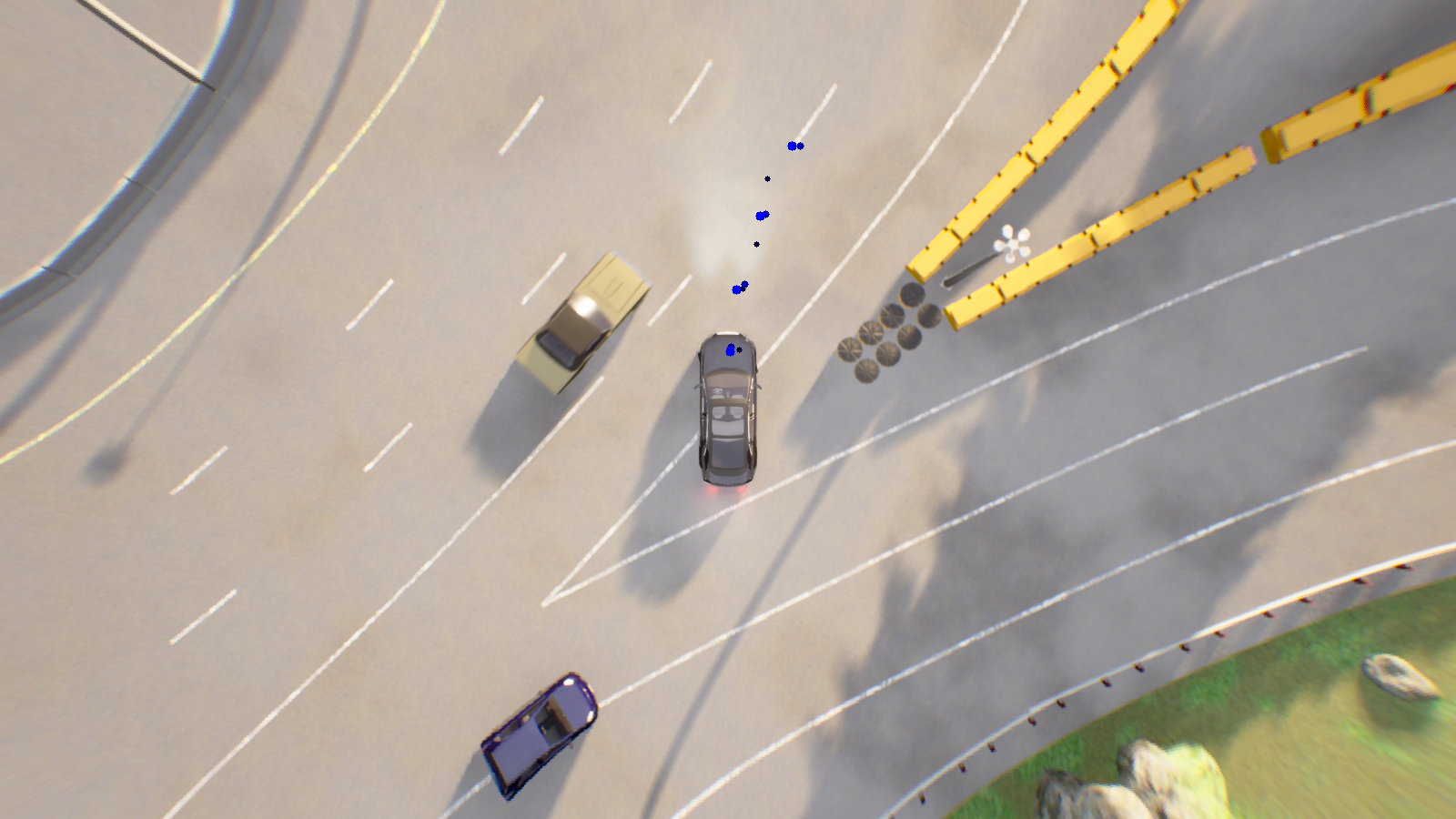}
		\captionsetup{width=8.5cm}
		\caption{\textbf{Merge}. The refined trajectory leaves more advance for the merging, which leads to a safer and smoother driving.}
		\end{subfigure}%
		\begin{subfigure}[t]{0.49\textwidth}
			\centering
			\includegraphics[width=8.5cm]{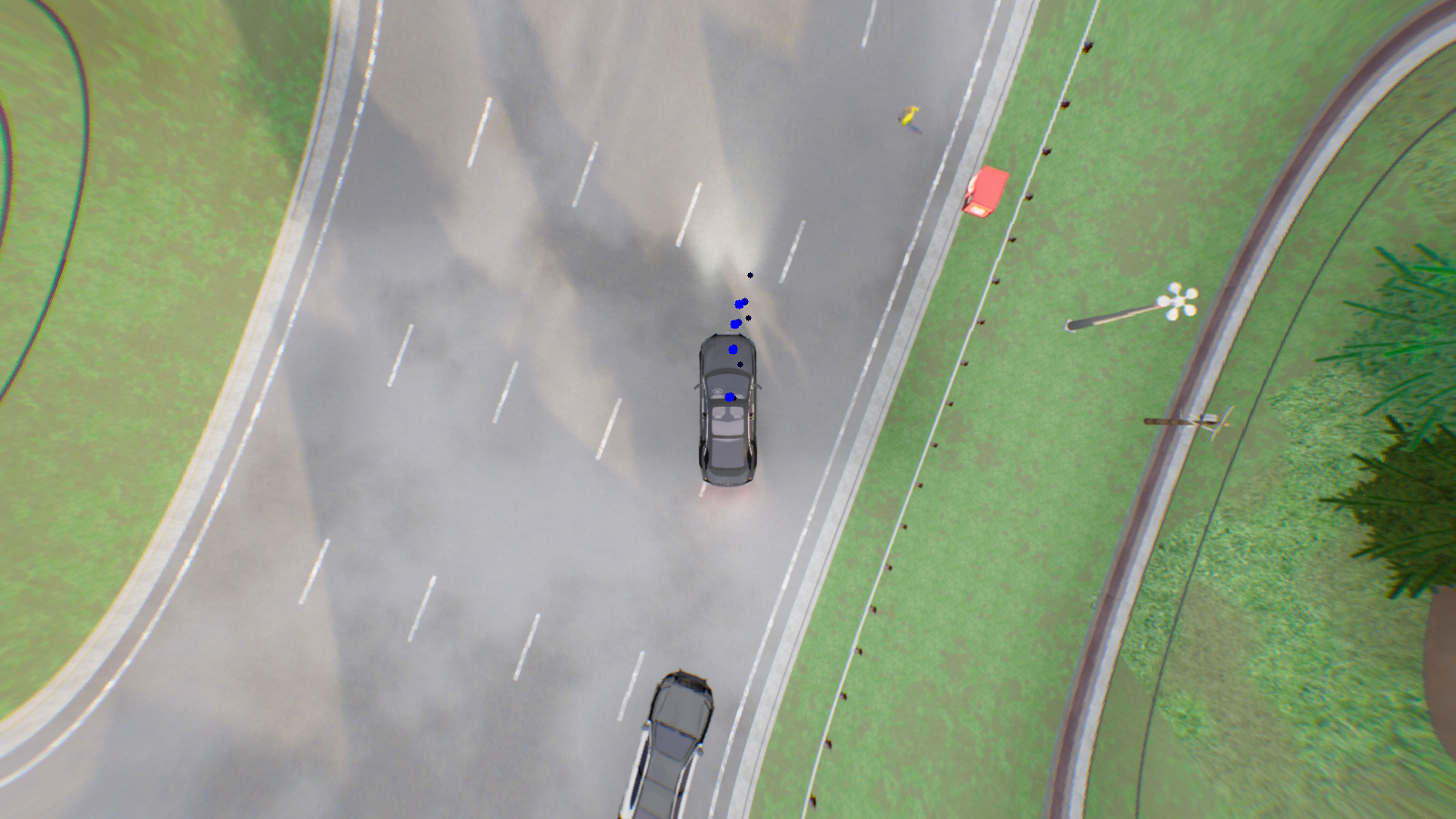}
			\captionsetup{width=8.5cm}
			\caption{\textbf{Lane Changing.} The refined trajectory notices the jay-walker and leads to a emergency stop during the lane-changing process.}
		\end{subfigure}%
		\caption{Visualization for the predictions from different layers of decoder. Larger and brighter dots are from deeper layers.}
\label{fig:vis}
\end{figure*}

\section{Implementation Details}
Since an end-to-end autonomous driving model is a large system, we provide details of our implementation so that it is easier for the community to reproduce. We will make the code and model publicly available.

\subsection{Data Collection}
We use Roach~\cite{zhang2021roach} as the expert with a collision detector for emergency stop similar to~\cite{wu2022trajectoryguided}. 
We set the following sensors: 

We set four cameras with field of view (FOV) $150^\circ$: Front (x=1.5, y=0.0, z=2.5, yaw=$0^\circ$), Left (x=0.0, y=-0.3,  z=2.5,yaw=$-90^\circ$), Right (x=0.0, y=0.3, z=2.5, yaw=$90^\circ$), Back(x=-1.6, y=0.0, z=2.5, yaw=$180^\circ$) where the aforementioned coordinate and angle are all in the ego coordinate system. The output of each camera is a 900x1600 RGB image. Since Carla~\cite{Dosovitskiy17} simulates the Brown-Conrady distortion~\cite{10.1093/mnras/79.5.384}, we estimate the distortion parameter with the code of~\cite{IBISCape22}. The estimated parameter for the distortion is (0.00888296, -0.00130899,  0.00012061, -0.00338673,  0.00028834) and we use the parameter to 
calibrate images before we feed them into the neural network. We also collected the depth and semantic segmentation label of images.

We set one Lidar with 64 channels, upper FOV $10^\circ$, lower FOV $-20^\circ$, and frequency 10Hz, following the official protocol. We set it at (x=0.0, y=0.0, z=2.5, yaw=$0^\circ$).

We set an IMU to estimate the yaw angle, acceleration, and angular velocity of the ego vehicle. We set a GPS to estimate current world coordinate of the ego vehicle and a speedometer to estimate current speed of the ego vehicle.

Following the official setting, we also save the target point which might be hundreds meters away as well high-level commands (keep straight, turn left, turn right, etc) provided by the protocol.

As for additional supervision signals, we save the value function, BEV feature maps of different resolution, the 1D feature as well as the control actions of Roach. 

We convert all raw data into the ego coordinate system.

\subsection{Models}

Our code is based on  OpenMMLab~\cite{mmcv} with Pytorch~\cite{paszke2019pytorch}, where we use their official implementation of backbones and cooresponding ImageNet pretrained weights if applicable. We use ResNet50~\cite{he2016deep} as the image backbone. For the extra data settings, we use ConvNext-base~\cite{liu2022convnet}. We use the PAFPN~\cite{liu2018path} to obtain the multi-scale image features. As for the LSS~\cite{philion2020lift} and depth module, we adopt the code from~\cite{li2022bevdepth}. For the semantic segmentation module, we use a U-Net~\cite{ronneberger2015u}-like structure. We downsample all images to 450x800 to save GPU memory. We use the BEV grid of 21x21 for low computational burden and set the scale as (Front=30.4m, Back=-8.0m, Left=-19.2m, Right=19.2m) which matches with the scale of Roach. We use 2 frames as the input while for the extra data settings we use 3 frames. For the Lidar model, we use the SECOND~\cite{yan2018second} implemented by mmdetection3D which consists of HardSimpleVFE, SparseEncoder, SECOND, and SECONDFPN. For the decoder, we have described details in the main text. For the extra data setting, we train 3 heads to further enlarge the capacity of the decoder.

\subsection{Hyper-Parameters}
We use AdamW~\cite{loshchilov2018decoupled} optimizer with the learning rate 1e-4, cosine learning rate decay, effective batch size 128, and weight decay 1e-7. We train the model for 60 epochs. For hidden dimensions, we use 256 at most places. For loss weights, we tune them to make sure that each loss is around 1 at the beginning of training.

\begin{table*}[!t]
\centering
\small
\scalebox{0.95}{
\begin{tabular}{lcccccc}
\toprule
Method & Encoder & Decoder      & Modality  &  \#Parameters      &   MACs (G) & GPU Memory          \\ \midrule
CILRS~\cite{codevilla2019exploring} & ResNet + Flatten  &   MLP     &  C1 & 23.4M    & 1.4  & 1507M \\
LBC~\cite{chen2020learning} & ResNet + Flatten  &   MLP     & C3 & 23.1M   & 5.4  & 1627M \\
Transfuser~\cite{Chitta2022PAMI} & Fusion via Transformer & GRU & C3L1 & 165.8M  & 34.9 & 2755M \\
Roach~\cite{zhang2021roach} & ResNet + Flatten & MLP      & C1 & 23.4M  & 17.1  & 2171M \\
LAV~\cite{chen2022lav}    & PointPaiting & Multi-layer GRUs     & C4L1  &  27.5M   & 45.5 & 2493M \\
TCP~\cite{wu2022trajectoryguided}   & ResNet + Flatten & GRU      & C1  &  25.8M & 17.1 & 2177M \\

MILE~\cite{hu2022model}  & ResNet + Flatten & GRU     & C1  & 54.1M   & 11.2  & 2485M \\
Interfuser~\cite{shao2022interfuser} & Fusion via Transformer & Transformer + GRU  & C3L1     & 82.8M & 46.5 & 1823M            \\
\textbf{ThinkTwice}    & Geometric Fusion in BEV &     Look-Predict-Refine &  C4L1  & 120.2M   & 1170+45  & 4019M \\
 \bottomrule
\end{tabular}}
\vspace{-3pt}
\caption{\textbf{Comparison of Computational Burden} MACs and \#Parameters are calculated by the Python package \textit{thop}. MACs and GPU memory is calculated under the inference mode of models. For ThinkTwice, the MACs is 1170G for the encoder (LSS module = 1157G) For \emph{Modality}, C denotes the camera sensor and L denotes the Lidar sensor. \vspace{-8pt} \label{tab:inference}}
\end{table*}

\subsection{Other Details}
For data augmentation which we only apply on images, we use the random color transformation similar to~\cite{wu2022trajectoryguided} and random crop before we project image features to the BEV grid. We stop the augmentation at the last ten epochs. The prediction time-horizon is 4. We apply gradient clip based on the L2 norm with the threshold of 35.

\section{Discussion about Inference Computation}
For the deployment of autonomous driving models, different from running on the cluster, they are usually running on edge devices with limited computational power and memory. Thus, it is important to discuss about the computation requirement and the memory footprint of different models during inference. Since few existing works give the computation related statistics in their papers, we run their official code to obtain an estimation. Note that all models are under their officially suggested environment, which means the version of packages such as Pytorch, Cuda, Carla, Numpy, OpenCV, etc may have some influence. All models are running under the same server with an RTX 3090 GPU. The estimation are in the Table~\ref{tab:inference}. We could observe that ThinkTwice has large MACs and GPU memory usage. It is because ThinkTwice is the only model adopts geometric fusion in BEV - specifically, LSS~\cite{philion2020lift} which requires 1153G MACs during the inference. We adopt it since the BEV representation inherently preserves the spatial relationships on the ground plane,
making it preferable for joint perception-planning and sensor fusion. From Table 4 in the main text, we could observe that simply combining the BEV representation with TCP~\cite{wu2022trajectoryguided} could achieve59 DS and the performance could be further enhanced to 65 DS with the proposed coarse-to-fine decoder. Actually, learning BEV representation is a heated topic in both industry and academia. To 
reduce the heavy computational burden of BEV-based model, more efficient implementations have been proposed in BEVFusion~\cite{liu2022bevfusion}, BEVDepth~\cite{li2022bevdepth}, BEVStereo~\cite{li2022bevstereo}, BEV-Pool-V2~\cite{huang2022bevpoolv2}, etc. Specifically designed edge devices and chips are also actively explored by the industry.

\section{Visualization}

We visualize different layers of predictions in the Fig.~\ref{fig:vis}. We can observe that with the future conditioned coarse-to-fine refinement, the driving process is safer and smoother.

\end{document}